\documentclass[manuscript,screen,review=false]{acmart}

\AtBeginDocument{%
  \providecommand\BibTeX{{%
    \normalfont B\kern-0.5em{\scshape i\kern-0.25em b}\kern-0.8em\TeX}}}

\usepackage{multirow}
\usepackage{makecell}
\usepackage{lineno}

\DeclareUnicodeCharacter{202F}{\,}

\setcopyright{acmcopyright}
\copyrightyear{2023}
\acmYear{2023}
\acmDOI{XXXXXXX.XXXXXXX}

\begin{document}
\title{Blockchained Federated Learning for Internet of Things: A Comprehensive Survey}

\author{Yanna Jiang}
\email{Yanna.Jiang@student.uts.edu.au}
\orcid{0000-0002-8176-6264}
\affiliation{%
  \institution{University of Technology Sydney}
  \city{Sydney}
  \state{NSW}
  \country{Australia}
  \postcode{2007}
}

\author{Baihe Ma}
\email{Baihe.Ma@uts.edu.au}
\orcid{0000-0003-4167-2797}
\affiliation{%
  \institution{University of Technology Sydney}
  \city{Sydney}
  \state{NSW}
  \country{Australia}
  \postcode{2007}
}

\author{Xu Wang}
\email{Xu.Wang-1@uts.edu.au}
\orcid{0000-0001-9439-6437}
\affiliation{%
  \institution{University of Technology Sydney}
  \city{Sydney}
  \state{NSW}
  \country{Australia}
  \postcode{2007}
}

\author{Ping Yu}
\authornote{Principal corresponding author}
\email{yuping0428@hit.edu.cn}
\orcid{0000-0002-8913-873X}
\affiliation{%
  \institution{Harbin Institute of Technology}
  \city{Harbin}
  \state{Heilongjiang}
  \postcode{150001}
  \country{China}
}

\author{Guangsheng Yu}
\email{Saber.Yu@data61.csiro.au}
\orcid{0000-0002-6111-1607}
\affiliation{%
  \institution{Data61, CSIRO}
  \city{Sydney}
  \state{NSW}
  \country{Australia}
  \postcode{2122}
}

\author{Zhe Wang}
\email{wz201018@163.com}
\affiliation{%
  \institution{Xidian University}
  \city{Xi'an}
  \state{Shaanxi}
  \postcode{710071}
  \country{China}
}

\author{Wei Ni}
\email{Wei.Ni@data61.csiro.au}
\orcid{0000-0003-0780-4637}
\affiliation{%
  \institution{Data61, CSIRO}
  \city{Sydney}
  \state{NSW}
  \country{Australia}
  \postcode{2122}
}

\author{Ren Ping Liu}
\email{RenPing.Liu@uts.edu.au}
\orcid{0000-0001-7001-6305}
\affiliation{%
  \institution{University of Technology Sydney}
  \city{Sydney}
  \state{NSW}
  \country{Australia}
  \postcode{2007}
}

\renewcommand{\shortauthors}{Yanna Jiang, et al.}

\begin{abstract}
The demand for intelligent industries and smart services based on big data is rising rapidly with the increasing digitization and intelligence of the modern world. 
This survey comprehensively reviews Blockchained Federated Learning (BlockFL) that joins the benefits of both Blockchain and Federated Learning to provide a secure and efficient solution for the demand. 
We compare the existing BlockFL models in four Internet-of-Things (IoT) application scenarios: Personal IoT (PIoT), Industrial IoT (IIoT), Internet of Vehicles (IoV), and Internet of Health Things (IoHT), with a focus on security and privacy, trust and reliability, efficiency, and data heterogeneity. 
Our analysis shows that the features of decentralization and transparency make BlockFL a secure and effective solution for distributed model training, while the overhead and compatibility still need further study.
It also reveals the unique challenges of each domain presents unique challenges, e.g., the requirement of accommodating dynamic environments in IoV and the high demands of identity and permission management in IoHT, in addition to some common challenges identified, such as privacy, resource constraints, and data heterogeneity.
Furthermore, we examine the existing technologies that can benefit BlockFL, thereby helping researchers and practitioners to make informed decisions about the selection and development of BlockFL for various IoT application scenarios.
\end{abstract}




\maketitle

\section{Introduction}
The Internet of Things (IoT), comprising smartphones, laptops, vehicles, and smartwatches, is ubiquitous and equipped with sensing and computing capabilities that enable accurate and effective data analysis and decision-making based on massive data and advanced models~\cite{12}.
Artificial Intelligence (AI) disciplines, especially the field of Machine Learning (ML), have been rapidly growing and widely applied to enhance the performance of these devices and drive the evolution of related industries~\cite{nguyen2020enabling,hu2021distributed}.
However, big-data-based applications bring significant risks and challenges, particularly in traditional centralized storage and computing approaches. 
The data collected by mobile devices and containing sensitive information is growing at an unprecedented rate, leading to a development bottleneck in the cloud-based data processing.

Various approaches have been proposed to meet the requirements of new-generation data storage, data processing, and privacy protection. 
One such approach is Federated Learning (FL), a distributed ML approach introduced in 2016 by McMahan et al.~\cite{1}. 
In the FL model, training data is kept locally on edge devices, instead of being uploaded to a central server. 
By only sharing the model parameters for aggregation, FL  mitigates the risk of privacy leakage during raw training data transmission, relieves the burden of centralized data storage and computation, and aligns well with the IoT development trend.

There is a growing focus on research in FL, recognizing the specific challenges and problems related to FL, such as heterogeneity and trust issues of the central server~\cite{li2020federated,yu2023ironforge}. 
To address these concerns and further advance development, Blockchain technology~\cite{2}, which enables safe data storage and sharing, was introduced as an alternative to classical FL's central server.
The integration of FL and Blockchain technology can leverage their strengths and enable the training of distributed models in a secure and decentralized way.
Despite the growing interest in FL and Blockchain, many existing studies have focused solely on FL~\cite{3} or Blockchain~\cite{4,wang2019survey}. 
FL and Blockchain have been considered simultaneously in work~\cite{5}, but the technical issues and challenges were presented separately for each area.

Researchers have recognized the potential for improved performance by integrating FL and Blockchain, so combining the two approaches has gained growing attention.
Nguyen et al.~\cite{6} presented a generic architecture based on FL and Blockchain, named FLchain, discussing their applications and challenges in edge computing. 
However, the authors did not address the crucial question of how to organize the content from these different fields comprehensively.
In contrast, Javed and collaborators~\cite{javed2022integration} primarily focused on applying the combination of FL and Blockchain to the Internet of Vehicles (IoV), with relatively less exploration of other potential application domains.
Similarly, the authors of~\cite{9739009} discussed the developments and challenges of Blockchain-based FL implementation on Unmanned Aerial Vehicles (UAVs), considering various application scenarios that can be combined in different directions. Their focus on a specific range of supported devices falls short of providing a comprehensive picture of the current state of technology combining FL and Blockchain.
The survey conducted by Zhu et al.~\cite{zhu2023blockchain} primarily categorized existing models that combine FL and Blockchain on integration methods. Although they mentioned various application directions, they did not delve into the commonalities and differences among these directions.

In this paper, we classify the state-of-the-art Blockchained FL (BlockFL) models that combine the technologies of FL and Blockchain technologies according to IoT application areas, based on which we further analyze the models according to the challenge tackled by these technologies. 
To provide a more organized and detailed discussion, we have categorized the broad IoT into four specific application areas:
\begin{itemize}
    \item Personal Internet of Things (PIoT): The PIoT focuses on improving the connectivity and automation of everyday objects, leveraging data generated by individual sensors and devices for a high degree of personalization, automation, and convenience~\cite{gupta2022personal}.
    \item Industrial Internet of Things (IIoT): The IIoT aims to produce intelligent manufacturing products and establish smart factories to enhance industrial processes and productivity~\cite{51}.
    \item Internet of Vehicles (IoV): The IoV directs attention to the segment of the IoT that centers on vehicles, dedicated to providing easy access to real-time road traffic information and high-quality in-vehicle services to improve the mobility and convenience of transportation~\cite{cheng2015routing}.
    \item Internet of Health Things (IoHT): The IoHT connects patients and healthcare facilities, using biomedical sensors for more efficient and convenient medical services~\cite{rodrigues2018enabling}.
\end{itemize}

As shown in Table~\ref{COMPARISON}, unlike prior works, our research concentrates on the technology combination in BlockFL, which utilizes Blockchain technology to securely store and manage local models and implement decentralized global model aggregation in FL, rather than the exploration of a single technology.
Furthermore, we focus on BlockFL models' performance and features in addressing various problems across the four diverse application domains mentioned above.
To the best of our knowledge, this paper covers most comprehensively the applications and techniques using a multilayered classification approach and presents recommendations for future research directions from multiple perspectives. 

\begin{table*}[]
    \centering
    \renewcommand\arraystretch{1.5}   
    \caption{Comparison of Related Work}
    \begin{tabular}{cccccccc}
    \toprule
    Reference  &  Year & Technique & Scenario & \makecell{Security \\ \& Privacy} & \makecell{Trusty \& \\ Reliability} & \makecell{Efficiency} & \makecell{ Data \\ Diversity } \\
    \midrule
    \cite{3} & 2021 & FL & PIoT & \checkmark  & \checkmark &  & \checkmark \cr
    \cite{4} & 2021 & Blockchain & IIoT, IoHT & \checkmark & \checkmark &  & \cr
    \cite{5} & 2021 & FL,Blockchain & PIoT, IoV & \checkmark &  & \checkmark &  \cr
    \cite{6} & 2021 & BlockFL & PIoT, IoV & \checkmark  &  & \checkmark & \checkmark \cr
    \cite{javed2022integration} & 2022 & BlockFL & IoV &  \checkmark &  &   & \cr
    \cite{9739009} & 2022 & BlockFL in UAVs & PIoT,IIoT, IoV, IoHT &  \checkmark &  &  \checkmark & \cr
    \cite{zhu2023blockchain} & 2023 & BlockFL & IIoT, IoV, IoHT & \checkmark &  &  & \checkmark \cr
    
    \multicolumn{2}{c}{\textbf{This paper}} & BlockFL & PIoT, IIoT, IoV, IoHT & \checkmark & \checkmark & \checkmark & \checkmark \\
    
    \bottomrule
    
    \end{tabular}

    \label{COMPARISON}
\end{table*}

The key contributions of this paper are summarized as follows.

\begin{itemize}
    \item We conduct a detailed analysis of BlockFL in four common scenarios, i.e., PIoT, IIoT, IoV, and IoHT, and highlight the challenges faced by BlockFL in these contexts. We also examine the advantages and disadvantages of BlockFL concerning these challenges comprehensively.
    
    \item We present an overview of the relationship between BlockFL, FL, and Blockchain, and perform a comparative classification of existing BlockFL applications and features in various scenarios, focusing on four essential aspects: security and privacy, trust and reliability, efficiency, and data heterogeneity.
    
    \item We analyze the common challenges and unique needs of BlockFL across different application domains and find that combining existing technologies (including cryptography, anomaly detection, compression techniques, and normalization) and enhancing the exploration of Blockchain components can drive the development of BlockFL. 
    
\end{itemize}

Our analysis reveals that features of decentralization and transparency make BlockFL a secure and effective solution for distributed model training, while the overhead and compatibility still need further investigation for the fruition of BlockFL.
Considering diverse application domains, our analysis also indicates that, besides the universal considerations of privacy protection, resource constraints and data heterogeneity, each domain presents unique challenges, e.g., the requirement of accommodating dynamic environments in IoV and the high demands of identity and permission management in IoHT.
It is anticipated that this paper can serve as an informative guide for future research efforts.

The rest of this paper is organized as follows. 
Section~\ref{Background} introduces the concepts and definitions of FL, Blockchain and BlockFL. 
Section~\ref{Applications} describes the different application scenarios of BlockFL. Section~\ref{Security and Privacy} -- \ref{Measurement Optimization} illustrate the latest application BlockFL models focusing on these different scenarios. 
The most prominent features of each reference are highlighted to show their advantages and limitations. 
Section~\ref{Challenges} summarizes the key lessons learned from the previous sections and puts forward future research directions. 
Finally, the conclusion of this paper and suggestions for the follow-up works are presented in Section~\ref{Conclusion}.


\section{Federated Learning, Blockchain, and BlockFL}
\label{Background}
This section introduces important concepts and models of FL and Blockchain, and analysis the basic framework of BlockFL that combines FL and Blockchain technologies.

\subsection{FL}
FL is a distributed ML framework~\cite{1} involving $N$ training participants and an aggregator. 
Participants, such as mobile devices, utilize their local datasets $\mathcal D$ for the training process and share their model parameters instead of their raw data. 
Meanwhile, the aggregator, such as a server, performs the aggregation operation on the shared parameters. 

\begin{figure}[!ht]
        \centering
       \includegraphics[width=0.6\columnwidth]{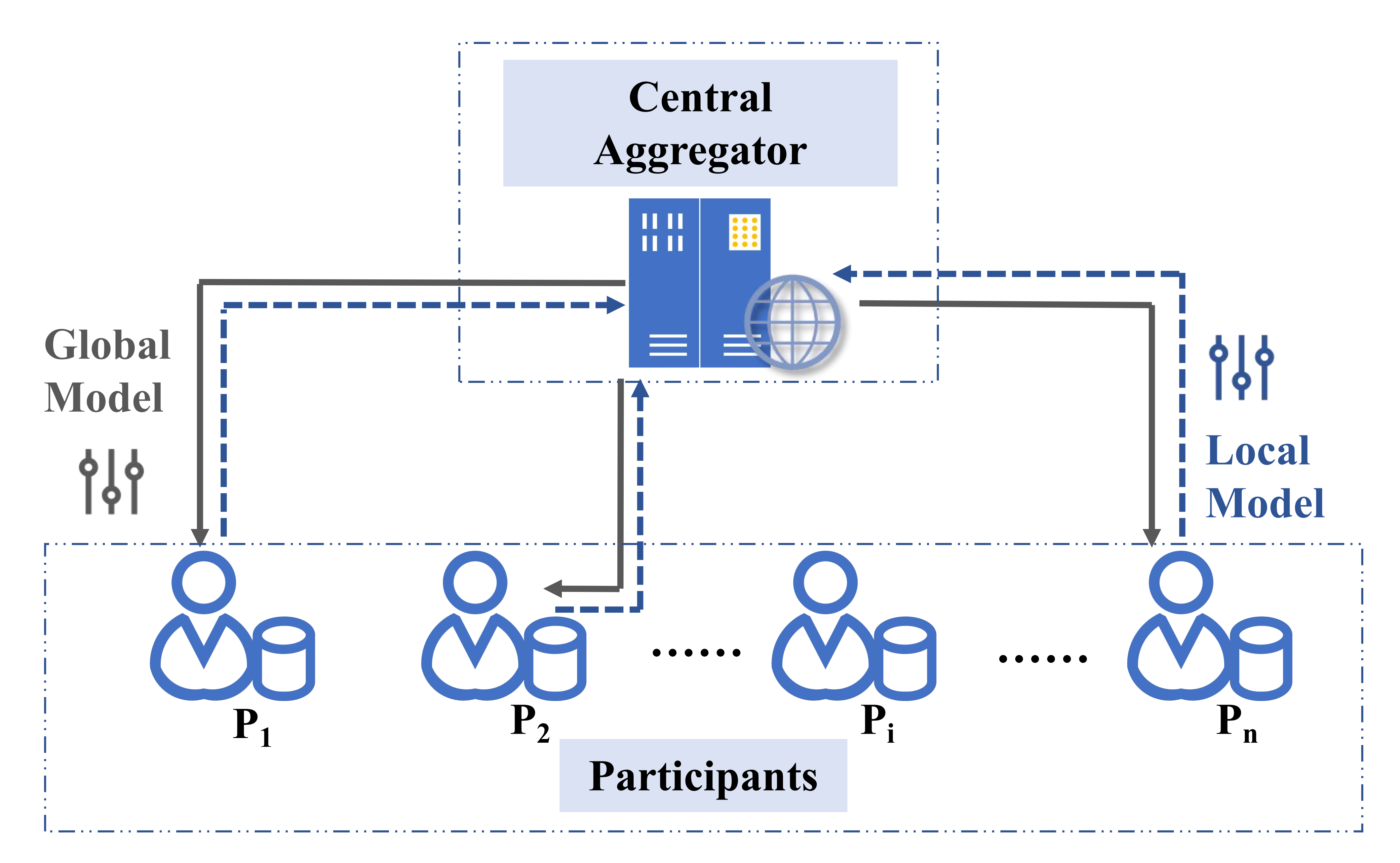}
        \caption{Traditional FL Process: Firstly, participants download the global model from the central aggregator. Then, participants perform local model training. Thirdly, participants upload their local models. Finally, the aggregator performs global model aggregation.}
        \label{fig_FL Process}
\end{figure}

In a typical FL process, there are four steps involved. 
By using $\{P_1, \ldots, P_i, \ldots, P_N\}$ to denote the $N$ training participants, the typical FL process is shown in the Fig.~\ref{fig_FL Process} and divided into the following parts:
\begin{itemize}
    \item First, the aggregator initializes the model and distributes it to all the participants;
    
    \item Next, the $i$-th participant $P_i$ trains the model using its local dataset $D_i$. The participant then obtains an improved model with an update $w_i$, which is achieved by minimizing a loss function $\mathcal F(w_i)$ as given by
    \begin{equation}
    w_i^* = \arg \min{\mathcal{F}(w_i)}, \quad i \in N,
    \end{equation}
    where the loss function $\mathcal F(w_i)$ is chosen differently depending on the FL algorithm to meet the model requirements of different scenarios;
    
    \item After local training, $P_i$ transmits the information of updated parameters to the aggregator for subsequent optimization;

    \item Finally, the aggregator calculates the shared parameters with the aggregation algorithm and updates the model according to the calculation results. 
\end{itemize}
Then, the updated model is returned to the participants, and the next round of training begins. These processes continue to loop until the model reaches the expected performance.

The model update in each loop is determined by the choice of the aggregation algorithm used in the FL process. One of the most commonly used aggregation algorithms is FedAvg~\cite{1}, which performs aggregation by computing the average during the FL process. Specifically, FedAvg calculates the shared parameters $w_G$ as follows:
\begin{equation}
w_G = \frac{1}{\sum_{i \in N} \lvert D_i \rvert} \sum_{i=1}^N \lvert D_i \rvert w_i,
\end{equation}
where $\lvert D_i \rvert$ represents the number of local training data in the dataset $D_i$ of participant $P_i$.

The FedAvg algorithm has limitations, such as the need to synchronize all updated parameters at each iteration and the consideration of dataset size in weight calculation. 
To address the limitations, several variants of FedAvg have been proposed to improve the effectiveness of aggregation. 
Reisizadeh et al.\cite{7} introduced FedPAQ, which allows for multiple local updates before sharing parameters and controls participant selection. 
Li et al.\cite{8} developed the FedProx algorithm, which uses a proximal term to reduce the computing consumption of heterogeneous data. 
Wang et al.~\cite{9} improved the FedMA algorithm, which applies a Bayesian non-parametric mechanism to adjust the model size based on distribution heterogeneity.

\begin{figure}[!ht]
        \centering
        \includegraphics[width=0.5\columnwidth]{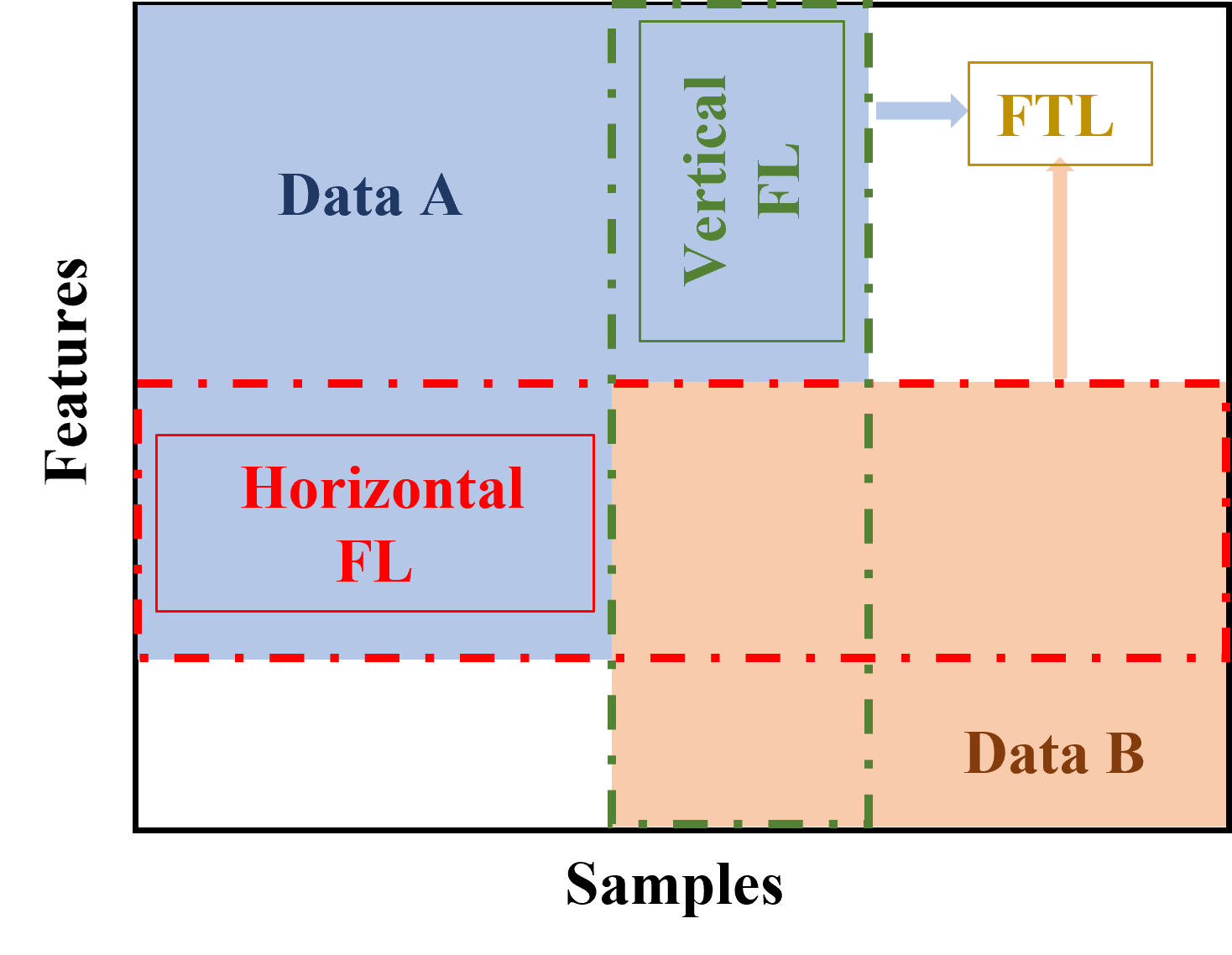}
        \caption{Categorization of FL: Horizontal FL, Vertical FL, and FTL. Horizontal FL is suitable for data with multiple consistent features, while Vertical FL is suitable for the same samples with different features. FTL is employed when the data features and samples are both overlapping less.}
        \label{fig_Categorization of FL}
\end{figure}

Based on the distribution characteristics of the data, FL can be categorized into horizontal FL, vertical FL, and Federated Transfer Learning (FTL)~\cite{yang2019federated}, as shown in Fig.~\ref{fig_Categorization of FL}. 
Horizontal FL is suitable when there are multiple consistent data characteristics with little overlap of samples in the participants' dataset. 
For example, the data from the same industry in different regions are very similar in characteristics, while the samples differ. 
In such a scenario, the model with horizontal FL can help deal with the data for analysis and judgment about the industry effectively and accurately.
Vertical FL, on the other hand, is used to enhance model capabilities by carrying out datasets with inconsistent characteristics but the same samples. 
This type of FL is useful when dealing with data from different industries in the same region.
FTL is employed when the data features and samples are both overlapping less. 
It can overcome insufficient data or labels and help obtain a more complex and accurate model.

\subsection{Blockchain}

As the underlying support of Bitcoin, Blockchain is a distributed ledger technology that uses cryptographic techniques to secure and maintain a decentralized database~\cite{10}.
Blockchain is designed to provide independent internal verification, communication, transmission, and storage while maintaining a reliable and transparent environment~\cite{yu2020enabling}.
This technique has the potential to meet various data requirements as it allows any peer to add new data and maintain synchronized information according to specific rules.

The Blockchain is structured as a series of blocks that store transactional information.
Each block is comprised of two parts: the header and the body. 
The header includes hash values of both the previous block and its own, enabling the blocks to link and form a continuous chain. 
In addition to hash values, the header stores essential information about the block, such as timestamps. 
The body of the Blockchain holds detailed information about transactions, where the first block in the Blockchain is typically known as the ``genesis'' block.
The features of Blockchain~\cite{monrat2019survey} have led to rapid development in existing industries, which can be described as follows:

\begin{itemize}
    \item Decentralization is the most significant feature of Blockchain. 
    With the consensus algorithm, Blockchain can verify and execute information transactions without requiring a trusted third party.
    \item Immutability is an essential trait of Blockchain, as all peers approve the information newly added through a decentralized consensus. 
    Hence, it is difficult and expensive to change the record of the Blockchain, which requires the consent of the majority.
    \item Auditability is also an important feature of Blockchain. 
    Each transaction in the Blockchain is accompanied by a unique hash and timestamp, and a copy of the Blockchain is held by all peers, allowing every peer to audit any specific transaction.
    \item Blockchain is autonomous. 
    With smart contracts, Blockchain can realize trust in physical machines, not bothered by anyone's interference.
\end{itemize}

\begin{table*}[!htbp]
\centering
\setlength\tabcolsep{2pt}
\renewcommand\arraystretch{1.5}   
\caption{Comparison of Different Blockchains}
\begin{tabular}{|c|c|c|c|c|c|}
    \hline
     \multirow{2}{*}{\textbf{Feature}} & \multirow{2}{*}{\textbf{Consensus Participant}} & \multirow{2}{*}{\textbf{Decentralization}} & \multirow{2}{*}{\textbf{Access}} & \multirow{2}{*}{\makecell{\textbf{Permission} \\ \textbf{Required}}} & \multirow{2}{*}{\makecell{\textbf{Transactions} \\\textbf{ Procession} } }\\
     & & & & & \\
     \hline
     Public Blockchains & All nodes & Complete & Public &  No & Slow \\
     \hline
     Private Blockchains & Single Organization & Partial & Controllable & Yes &  Fast \\
     \hline
     Consortium Blockchains & Multiple Organizations & Partial & Controllable & Yes &  Fast \\
     \hline
\end{tabular}
\label{Blockchains}
\end{table*}

Blockchains have already demonstrated their usefulness in the context of IoT~\cite{makhdoom2019blockchain,yublockchain,makhdoom2018blockchain}, and the capacity of blockchains has been analyzed in \cite{wang2019capacity,wang2021capacity,yu2020scaling} for IoT applications.
According to the application scenario, the Blockchain can be classified into three types~\cite{huo2022comprehensive}, as follows.
\begin{itemize}
    \item Public Blockchain:
    In the public Blockchain, all nodes participate in the consensus process and have the right to publish new blocks and access the whole Blockchain. 
    The public Blockchain is the most completely decentralized, and most of the familiar Blockchain entities belong to this category, such as Bitcoin and Ether.
    
    \item Private Blockchain:
    The nodes in a private Blockchain need permission to join the network and participate in Blockchain activities. 
    This type of Blockchain is suitable and often used for a single organization or enterprise, which has control over the consensus process, and thus, private Blockchain is not truly decentralized. 
    Compared with public Blockchain, private Blockchain is generally smaller in scale and controllable in access, making transactions faster to process and the system easier to implement. 
    
    \item Consortium Blockchain:
    Consortium Blockchain is based on the private Blockchain and built a consortium network across multiple organizations. 
    Permission is also necessary for the nodes in the consortium Blockchain to become members of the Blockchain. 
    The scale of the consortium Blockchain can be larger and involves more participating nodes than that of the private Blockchain, but in other performance characteristics, it is still consistent with the private Blockchain. 
\end{itemize}

A detailed comparison between the three Blockchain types is shown in Table~\ref{Blockchains}. 
In terms of the consensus process, all nodes of a public Blockchain can participate, while the consensus of a private Blockchain is controlled by a single organization. A consortium blockchain expands on the private Blockchain to include multiple organizations. 
Correspondingly, a public Blockchain has complete decentralization, and its access is public without requiring permission. By contrast, private and consortium Blockchains, on the other hand, are only partially decentralized and more controlled, where nodes need permission to access.

\subsection{BlockFL}

In this section, we illustrate a BlockFL model that combines the technology of FL and Blockchain. 
The BlockFL model uses Blockchain as a reliable model storage and an aggregator to replace the central server in FL for decentralization.
Table~\ref{DIFFERENCE} presents a comparison between the features, advantages, and disadvantages of FL, Blockchain and BlockFL. 
The comparison reveals that BlockFL successfully integrates the strengths of FL and Blockchain, while mitigating some of their limitations, making it a promising solution for various applications.

\begin{table*}[!htb]
    \centering
    \setlength\tabcolsep{4pt}
    \renewcommand\arraystretch{0.9}   
    \caption{Difference between FL, Blockchain and BlockFL}
    \begin{tabular}{|c|c|c|c|}
    \hline
     \multirow{2}{*}{\textbf{Technique}} & \multirow{2}{*}{\textbf{Feature}} & \multirow{2}{*}{\textbf{Advantage}} & \multirow{2}{*}{\textbf{Limitation}} \\
     & & & \\
    \hline
       \multirow{5}{*}{FL}  & \multirow{5}{5cm}{Privacy-preserving Model Training} & \multirow{5}{*}{\makecell{Data Privacy, \\ Reduced Data Transmission, \\ Improved Model Accuracy.} } &  \multirow{5}{*}{\makecell{ Centralization, \\ Dependent on \\ Participants' Performance, \\ Potential Biases. } }\\
       & & & \\
       & & & \\
       & & & \\
       & & & \\
    \hline
       \multirow{4}{*}{Blockchain}  & \multirow{4}{5cm}{Secure and Transparent Data Sharing and Storage} & \multirow{4}{*}{\makecell{ Decentralization, \\ Immutability \& Auditability, \\ Autonomous}} & \multirow{4}{*}{\makecell{Scalability issues, \\ High Energy Consumption } }\\
       & & & \\
       & & & \\
       & & & \\
    \hline
       \multirow{5}{*}{BlockFL}  & \multirow{5}{5cm}{Multi-party Decentralized and Secure Collaborative Model Training} & \multirow{5}{*}{\makecell{Decentralization, \\ Data Security, \\ Privacy Preservation, \\  Transparency } } &  \multirow{5}{*}{ \makecell{ Increased Communication \\ \& Computation Overhead,\\ Limitations of Network \\ \& Device Compatibility}}\\
       & & & \\
       & & & \\
       & & & \\
       & & & \\
    \hline
    \end{tabular}
    \label{DIFFERENCE}
\end{table*}

The BlockFL system model~\cite{kim2019blockchained} consists of two parts: the local learning process (running on mobile devices) and the integrated calculation process (implemented on the Blockchain). 
In the BlockFL system, there are two main actors: participants (i.e., mobile devices) who use their local datasets to learn preliminary models, and miners in the Blockchain who verify models and facilitate aggregate calculations. 
The participants and miners can be either the same or different entities.
The BlockFL system can be described as follows.

Once an expected model is requested, a crowdsourcing task is created on the Blockchain. 
Interested participants begin the local learning process by downloading the initial model from the Blockchain and training their local model with their respective datasets. 
The progress of the training process depends on the factors, such as the amount of data and computing power available to the participants. 
With multiple training rounds, participants can get their local models that achieve high performance on their local datasets. 
Then, participants sign the hash values of their models with their private keys and send their models to the Blockchain for privacy protection and security, which is different from the traditional FL process.

Then, the BlockFL system operates within the Blockchain, which serves as permanent and immutable storage for machine learning models. 
Transactions processed by the miners include verifying the submitted model's related signatures and scoring its contribution. 
In verification, the miners are responsible for rejecting the fake data from adversaries in the submitted models.
The score of the model is a comprehensive parameter that considers both the accuracy of the model and the size of the training dataset. 
The score affects the rewards of participants who submit the model and is used to determine the weight of the global aggregation. 
The miners compete to be the leader who generates a new block for the integrated global model. 
The elected leader calculates the global model parameters based on the submitted local models and their corresponding scores. 
The leader creates a new block consisting of the calculated global model and other participants' signatures and agreement. 
The consensus protocol is based on Byzantine Fault Tolerance (BFT)~\cite{castro1999practical}, which ensures the system's security by assuming that more than $2/3$ of miners are trustworthy.

\begin{figure}[!htb]
        \centering
       \includegraphics[width=0.6\columnwidth]{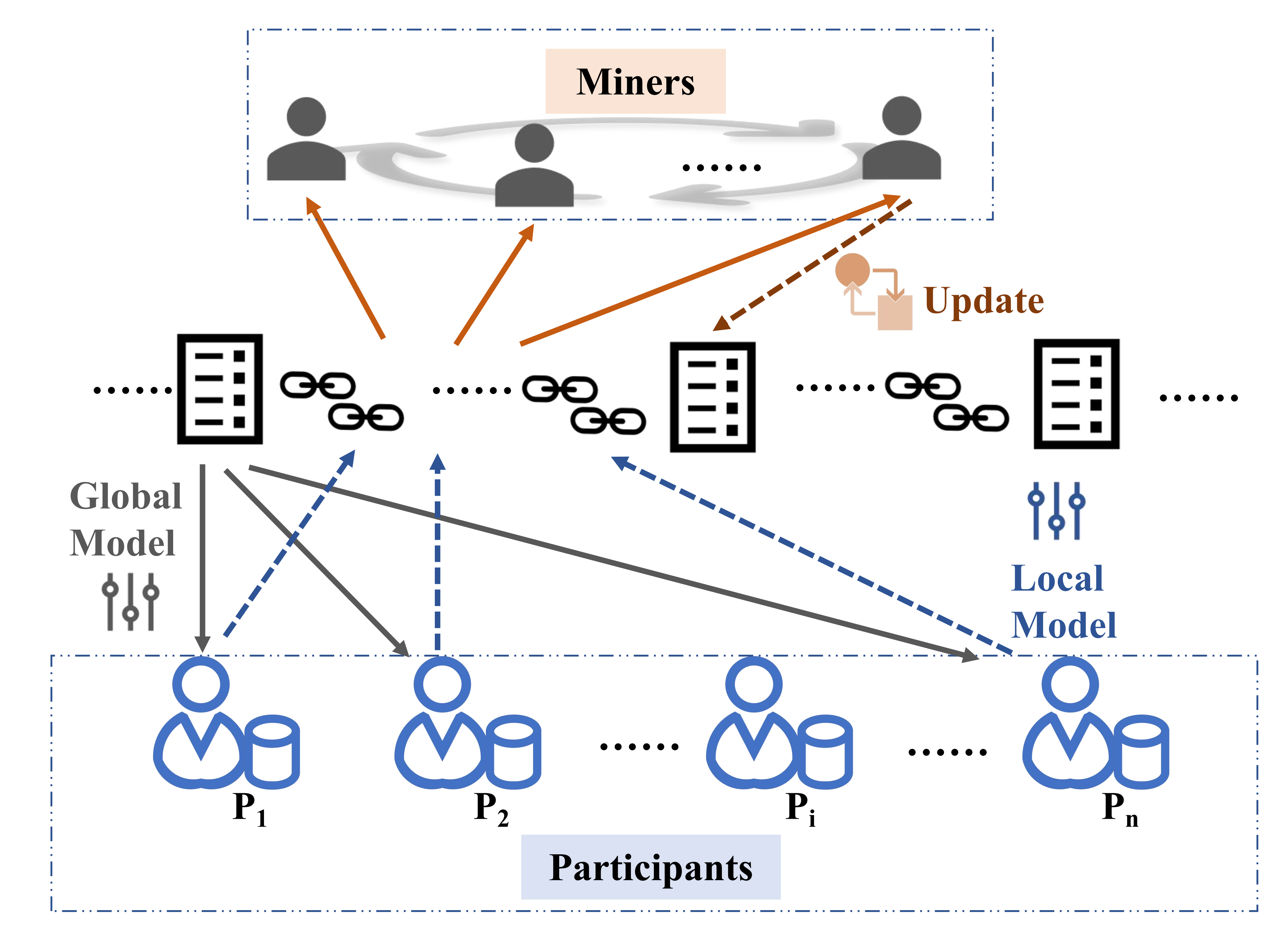}
        \caption{BlockFL Process: First, participants download the global model from Blockchain; Then, participants perform local model training. Thirdly, participants upload their local models to Blaockchain; Next, miners process transactions to verify and score; After that, miners conduct a leadership competition; Finally, the leader elected generates a new block with the updated global model.}
        \label{fig_BlockFL Process}
\end{figure}

\begin{table*}[]
\setlength\tabcolsep{1.5pt}  
    \centering
    \renewcommand\arraystretch{0.98}   
    \caption{Comparison of Technical Features of BlockFL Models}
    \begin{tabular}{|c|c|c|c|c|c|c|}
    \hline
     \multirow{3}{*}{\makecell{\textbf{BlockFL} \\ \textbf{Models}}} & \multirow{3}{*}{\makecell{\textbf{Synchro-}\\\textbf{nization}}} & \multirow{3}{*}{\makecell{\textbf{Chain} \\\textbf{Structure}}} & \multirow{3}{*}{\textbf{Permission}} & \multirow{3}{*}{\makecell{\textbf{Consensus}\\\textbf{Mechanism}}} & 
     \multirow{3}{*}{\makecell{\textbf{Appli-}\\\textbf{cation}}} & \multirow{3}{*}{\textbf{Features}} \\
      & & & & & & \\
      & & & & & & \\
    \hline
    \multirow{4}{*}{\makecell{Autonomous \\ BFL \\ \cite{70}}} & \multirow{4}{*}{\makecell{Synch-\\ronous}} & \multirow{4}{*}{Blockchain} & \multirow{4}{*}{Public} & \multirow{4}{*}{PoW} & 
    \multirow{4}{*}{IoV} & \multirow{4}{*}{\makecell[l]{End-to-end trustworthiness assurance; \quad \ \\Delay minimization and block arrival \\ rate optimization.}}\\
    & & & & & & \\
    & & & & & & \\
    & & & & & & \\
    \hline
     \multirow{4}{*}{\makecell{BAFL\\ \cite{35}}}& \multirow{4}{*}{\makecell{Asynch-\\ronous}} & \multirow{4}{*}{Blockchain} & \multirow{4}{*}{Public} & \multirow{4}{*}{PoW} & 
     \multirow{4}{*}{PIoT} & \multirow{4}{*}{\makecell[l]{Faster convergence of the global model; \\Score for secure evaluation;\\ Dual strategy trade-off parameters.}}\\
     & & & & & & \\
     & & & & & & \\
     & & & & & & \\
    \hline
    \multirow{5}{*}{\makecell{ChainsFL \\ \cite{36}}} & \multirow{5}{*}{\makecell{Synch-\\ronous +\\Asynch-\\ronous}} & \multirow{5}{*}{\makecell{Blockchain \\ + DAG}} & \multirow{5}{*}{Public} & \multirow{5}{*}{\makecell{Raft +\\Tangle\\Consensus}} & 
    \multirow{5}{*}{PIoT} & \multirow{5}{*}{\makecell[l]{Two-layer Blockchain for security and \\ scalability enhancement;\\ Synchronous and asynchronous training \\combination for efficiency improvement.}}\\
    & & & & & & \\
    & & & & & & \\
    & & & & & & \\
    & & & & & & \\
    \hline
    \multirow{4}{*}{\makecell{FedAC\\ \cite{32}}} & \multirow{4}{*}{\makecell{Asynch-\\ronous}} & \multirow{4}{*}{Blockchain} & \multirow{4}{*}{Public} & \multirow{4}{*}{PoW} & 
    \multirow{4}{*}{PIoT} & \multirow{4}{*}{\makecell[l]{Considering a staleness coefficient;\quad \quad \ \\Avoidance of single-point failures and \\secure protection for cyberattacks.}}\\
    & & & & & & \\
    & & & & & & \\
    & & & & & & \\
    \hline
    \multirow{6}{*}{\makecell{FL-Block\\ \cite{63}}} & \multirow{6}{*}{\makecell{Synch-\\ronous}} & \multirow{6}{*}{Blockchain} & \multirow{6}{*}{Public} & \multirow{6}{*}{PoW} & 
    \multirow{6}{*}{IIoT} & \multirow{6}{*}{\makecell[l]{Only global updates pointer saved \\ on-chain for block generation efficiency;\\Prevention of single point failure;\\Poisoning attacks proved to be eliminated;\\ Optimal block generation rate analysis.}}\\
    & & & & & & \\
    & & & & & & \\
    & & & & & & \\
    & & & & & & \\
    & & & & & & \\
    \hline
    \multirow{5}{*}{\makecell{Hierarchical\\ BlockFL\\ \cite{76}}} & \multirow{5}{*}{\makecell{Synch-\\ronous}} & \multirow{5}{*}{\makecell{Hierarchical\\Blockchain }} & \multirow{5}{*}{Public }& \multirow{5}{*}{PoK} & 
    \multirow{5}{*}{IoV} & \multirow{5}{*}{\makecell[l]{Knowledge sharing framework with one \\top chain and multiple ground chains;\\ Hierarchical FL algorithm with a bottom \\ knowledge aggregation middle layer.}} \\
    & & & & & & \\
    & & & & & & \\
    & & & & & & \\
    & & & & & & \\
    \hline
    \multirow{5}{*}{\makecell{MAS \\BlockFL \\ \cite{81}}} & \multirow{5}{*}{\makecell{Synch-\\ronous}} & \multirow{5}{*}{Blockchain} & \multirow{5}{*}{Authorized} & \multirow{5}{*}{PoW} & 
    \multirow{5}{*}{IoHT} & \multirow{5}{*}{\makecell[l]{Parallel training of IoHT classifiers;\\ Private Blockchain for secure data sharing\\ and privacy protection;\\Allow specific tasks assigned to agents.}}\\
    & & & & & & \\
    & & & & & & \\
    & & & & & & \\
    & & & & & & \\
    \hline
    \multirow{4}{*}{\makecell{PermiDAG\\ \cite{72}}} & \multirow{4}{*}{\makecell{Asynch-\\ronous}} & \multirow{4}{*}{\makecell{Blockchain \\ + DAG}} & \multirow{4}{*}{Authorized} & \multirow{4}{*}{\makecell{ DPoS + \\Simplified \\PoW} }& 
    \multirow{4}{*}{IoV} & \multirow{4}{*}{\makecell[l]{Hybrid scheme for efficiency;\\ DRL algorithm for participant selection;\\Two-stage quality verification.} }\\
    & & & & & & \\
    & & & & & & \\
    & & & & & & \\
    \hline
    \multirow{5}{*}{\makecell{Secure Data\\Sharing\\ Scheme\\ \cite{56}}} & \multirow{5}{*}{\makecell{Synch-\\ronous}} & \multirow{5}{*}{Blockchain} & \multirow{5}{*}{Authorized} & \multirow{5}{*}{PoQ} & 
    \multirow{5}{*}{IIoT} & \multirow{5}{*}{\makecell[l]{Permissioned Blockchain for data sharing;\\Integration of differential privacy to FL;\\Improved computing resources utilization\\ and data sharing efficiency.}}\\
    & & & & & & \\
    & & & & & & \\
    & & & & & & \\
    & & & & & & \\
    \hline
     \multirow{4}{*}{\makecell{ VFChain \\ \cite{26}}} &  \multirow{4}{*}{\makecell{Synch-\\ronous}} &  \multirow{4}{*}{Blockchain} &  \multirow{4}{*}{Public} &  \multirow{4}{*}{PBFT} &  
     \multirow{4}{*}{PIoT} &  \multirow{4}{*}{\makecell[l]{Committee for verifiable proofs;\\ DSC for effective data authentication;\\ Multiple-Model tasks DSC optimization.}} \\
    & & & & & & \\
    & & & & & & \\
    & & & & & & \\
    \hline
    \end{tabular}
    \label{technical paper}
\end{table*}

The global model in the Blockchain is regularly updated, prompting participants to download and train it with their local datasets repeatedly. The iterative process of local learning and integrated calculation continues until the global model researches the expected level of accuracy and convergence.


The process of a BlockFL model is shown in Fig.~\ref{fig_BlockFL Process}.
Compared to traditional FL, BlockFL introduces a more complex process by adding miner validation and leader election, leveraging Blockchain technology to replace the role of traditional aggregators. The uploaded and downloaded global models in BlockFL are stored in secure blocks, and model aggregations are completed through miner campaigns. This eliminates the dependence on unreliable aggregators in FL, reducing associated risks and improving the security and trustworthiness of the overall process.

Table~\ref{technical paper} provides a comprehensive summary of the technical features of various BlockFL models, analyzing and comparing them based on factors such as training synchronization, chain structure, consensus mechanisms, and permission. This analysis offers valuable insights into the current state of development of popular BlockFL models.


\section{Applications of BlockFL in IoT}
\label{Applications}

\begin{figure}[!ht]
        \centering
        \includegraphics[width=0.6\columnwidth]{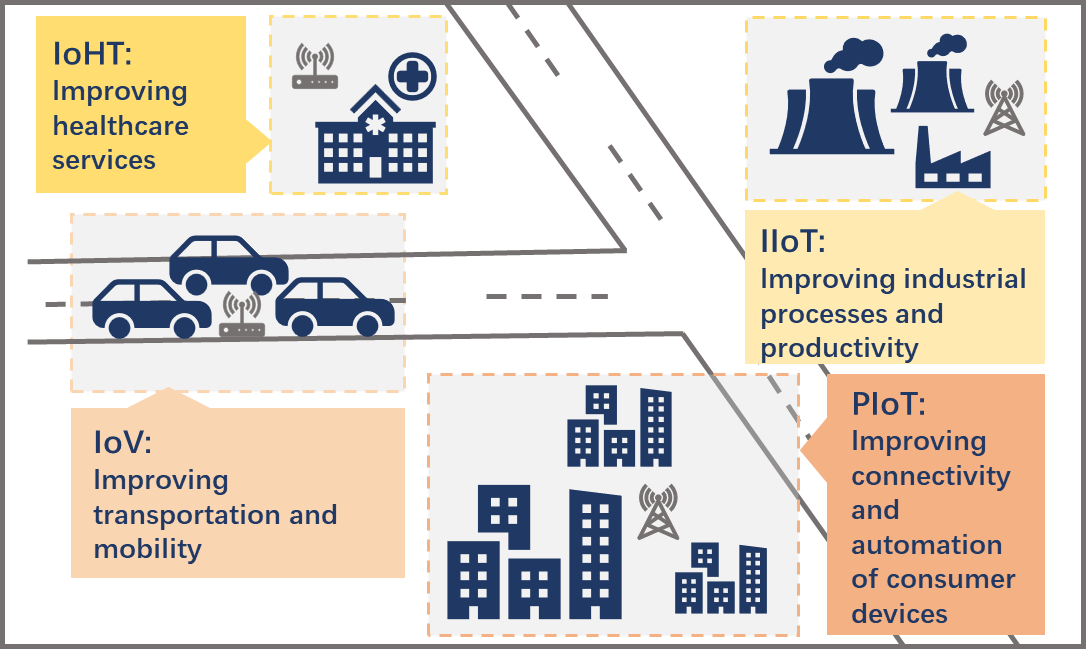}
        \caption{Application scenarios of BlockFL: PIoT, IIoT, IoV, and IoHT. PIoT focuses on improving connectivity and automation of everyday objects, while IIoT focuses on industrialized processes and productivity improvements. IoV focuses on transportation, and IoHT focuses on medical health. All aspects of life and work are encompassed by these four areas.}
        \label{fig_application}
\end{figure}

The BlockFL framework has been widely implemented in various scenarios to enhance security, privacy, reliability, and efficiency. 
We analyze the key indicators and challenges based on the development of BlockFL in different application domains. 
Existing studies on the integration of FL and Blockchain are divided into four parts based on their application scenarios: PIoT, IIoT, IoV, and IoHT, as shown in Fig.~\ref{fig_application}. 

\subsection{PIoT}

The IoT refers to a vast network of heterogeneous physical objects interconnected through the internet, which are capable of collecting and sharing information.
PIoT, a sub-domain of IoT, aims to enhance convenience and efficiency for individuals by improving the connectivity and automation of everyday objects, including home appliances and wearable devices. This focus on consumer-level applications distinguishes PIoT from the industrial applications of IIoT.

The underlying technologies of PIoT enable the objects to coordinate decisions and provide smart services. 
With the rapid development of technologies, an increasing number of devices would participate in the PIoT ecosystem. 
According to statistical research, there will be over 75.44 billion PIoT devices worldwide by 2025, generating more than 79.4 zettabytes of data~\cite{rizvi2020threat}, and the economic impact is expected to be 2.7 to 6.2 trillion dollars~\cite{manyika2013disruptive}.
Therefore, the development and optimization of the PIoT have become of paramount importance.

The basic model of PIoT is a 5-layer architecture~\cite{11}, including the object layer, object abstraction layer, service management layer, application layer, and business layer. 
Based on the elements, four challenges need to be solved urgently with the BlockFL model in PIoT to meet the growing demand: 

\begin{itemize}
    \item Security and Privacy.
     The vast amounts of data generated by PIoT devices and trained models during PIoT applications require robust privacy protection measures to prevent unauthorized access or malicious use, particularly in the context of data generated during PIoT's sensing, communication, and computation processes.
    \item Reliability.
    For the PIoT to be practically applied, it is crucial that the information and services it provides are trusted and reliable. 
    \item Efficiency.
    Efficient and effective PIoT models are essential to enable the practical application of PIoT systems, particularly in resource-limited environments.
    \item Incentive.
    Optimizing PIoT models rely on large amounts of diverse data, which requires the development of a reasonable incentive mechanism that encourages device participation. 
\end{itemize}

\subsection{IIoT}

IoT technologies are expected to provide promising solutions in various industries, such as mining production, firefighting, agriculture, and the food industry~\cite{48}. 
As a result, IIoT, the application of IoT in industries, has garnered a lot of attention. 
The formation of IIoT marks the beginning of Industry 4.0~\cite{49}, where industrial production and network connectivity are combined for intelligence and visualization. 
The IIoT architecture consists of machines, networks, clouds, and applications~\cite{50}, where the machines are equipped with various sensors and network connection functions. 

Compared with other areas, IIoT has specific features and requirements~\cite{52}. The core feature of IIoT is to serve industrial productions, so IIoT pursues automation in the production process. The technologies of IIoT help to make intelligent decisions by connecting all machines and equipment, data, process, operators and decision-makers from the factory to the office. Based on the characteristics, there are three issues in the application of IIoT:

\begin{itemize}
    \item Security.
    The development of IIoT has led to exponential growth in data generated by various types of equipment. With the value of data containing important information, concerns about data security in IIoT have arisen.
    \item Stability.
    Reliability and stability are essential for the practical application of IIoT models. The IIoT model must function effectively in specific working environments and withstand varying degrees of disruption or attack.
    \item Resource limitations.
    During research and analysis, researchers often assume that the devices participating in model training have unlimited energy and computing power. However, in reality, the industrial machines used in manufacturing may not meet the requirements of the theoretically optimal model. Due to cost considerations, various performance aspects of industrial equipment often have limitations such as capacity, energy, and communication ability.
   
\end{itemize}

\subsection{IoV}

Intelligent vehicles are rapidly advancing with the assistance of automotive devices that enhance sensing, computation, and communication capabilities~\cite{cui2022vehicular}. 
Tens of thousands of vehicles involuntarily connect with each other, forming vehicular networks that communicate and share information via Vehicle-to-Vehicle (V2V) and Vehicle-to-Infrastructure (V2I)~\cite{67}. 
Similar to the PIoT, the convergence of vehicular networks and Internet technology gives rise to IoV, with Tesla Motors being one of the globally renowned examples.

The structure of IoV can be divided into three layers: the perception layer, networking layer, and application layer~\cite{68}. 
At the perception layer, vehicles in IoV can sense the surrounding environment through sensors, such as radar and cameras, assisting in driving in various situations. 
The networking layer facilitates communication among vehicles and infrastructure, allowing for the real-time exchange of information on traffic conditions and road hazards via Vehicle-to-Vehicle (V2V) and Vehicle-to-Infrastructure (V2I) communication. 
At the application layer, devices with high computing and storage capabilities are equipped in IoV, enabling information sharing and model learning. 
These capabilities facilitate a wide range of applications in IoV, such as autonomous driving, intelligent transportation, and traffic management systems.

However, the comprehensiveness of the IoV makes its development more challenging than the development of other IoT domains~\cite{69}. For instance, autonomous driving requires the cooperation of vehicle parts, the allocation of resources, the collection of information, and the analysis of decisions; intelligent transport systems in smart cities need to coordinate more resources and big data and consider more possibilities in different scenarios. In terms of current research works, there are four issues to be solved in the applications of the BlockFL model in the IoV:

\begin{itemize}
    \item Privacy.
    In order to learn a practical IoV model, a large number of various types of data needs to be shared, including sensitive private information such as frequently visited addresses, real-time road conditions, driving routes, driving preferences, and more.
    \item Reliability.
    The high mobility of vehicles in the IoV introduces dynamic environments and constantly changing circumstances, which demands high reliability and stability of the models. 
    \item Timeliness.
     Additionally, the information in the IoV changes rapidly, making the timeliness of decisions crucial for both autonomous driving and intelligent transportation systems. Vehicles must respond quickly to real-world situations.
    \item Resource limitations.
    The dynamic nature of vehicular environments also presents a challenge for resource allocation. Despite advancements in in-vehicle computing and communication technologies, there is still a gap to reach the optimal solution for model learning in theory.
\end{itemize}

\subsection{IoHT}

Wearable devices, including smartwatches and fitness trackers, have become increasingly popular.
The devices provide continuous physiological data for healthcare, contributing to the development of IoHT.
The IoHT is an extension of the PIoT and focuses on collecting and analyzing health-related data for monitoring and diagnosing health conditions. 
IoHT has a wide range of applications, including intelligent healthcare, telemedicine, home care, and fitness monitoring~\cite{78}.

The IoHT has gained increased attention since the outbreak of COVID-19, as intelligent healthcare with IoHT technologies plays an essential role in controlling the spread of infectious diseases using biomedical sensors and health data systems~\cite{79}. 
By using corresponding technology, various real-time physiological data, e.g., heart rate, body temperature, blood pressure, and motion trajectory, are collected from people who use IoHT devices, which can be analyzed to predict and control the spread of infectious diseases.

As an emerging industry, the existing researches on IoHT technologies are not comprehensive and in-depth. But the IoHT has shown great potential in the intelligentization of medical care and the personalization of healthcare, especially in the face of a large-scale worldwide epidemic like COVID-19. In the development process of the IoHT, there are two prominent issues worth noting:

\begin{itemize}
    \item Privacy.
    As mentioned, IoHT involves a vast amount of sensitive personal physiological information, the leakage of which can pose significant privacy risks. Therefore, ensuring privacy protection and data security in the IoHT is crucial. 
    \item Heterogeneity.
    Due to the diversity of medical equipment and the complexity of application scenarios, it seems impossible to enforce the structure for all data in IoHT with a harmonized format. So handling heterogeneous data and analytical modeling becomes critical for IoHT model optimization.
\end{itemize}

\begin{table*}[!ht]
    \centering
    \caption{BlockFL in different application scenarios}
    \begin{tabular}{|c|c|c|}
    \hline
    \multirow{3}{*}{\textbf{Scenario}} & \multirow{3}{*}{\textbf{Features}} & \multirow{3}{*}{\makecell[l]{\textbf{Challenges that Can}\\\textbf{Benefit from BlockFL}} } \\
    & & \\
    & & \\
    \hline
       \multirow{5}{*}{PIoT} & \multirow{5}{*}{\makecell[l]{Collection and Analysis of vast amounts of consumer data; \\Improving decision-making and operational efficiency}} & \multirow{5}{*}{\makecell{Security \& Privacy, \\ Reliability, \\ Efficiency, \\ Incentive} } \\
    & & \\
    & & \\
    & & \\
    & & \\
    \hline
       \multirow{4}{*}{IIoT} & \multirow{4}{*}{\makecell[l]{Real-time monitoring and control of industrial processes;\\ Improving productivity and reducing downtime}} & \multirow{4}{*}{\makecell{Security, \\ Stability, \\ Resource Limitations}} \\
    & & \\
    & & \\
    & & \\
    \hline
       \multirow{5}{*}{IoV} & \multirow{5}{*}{\makecell[l]{Enhances road safety; Reduces traffic congestion; \\ Provides a more comfortable driving experience}} & \multirow{5}{*}{\makecell{Privacy,  \\ Reliability, \\ Timeliness, \\ Resource Limitations} }\\
    & & \\
    & & \\
    & & \\
    & & \\
    \hline
       \multirow{4}{*}{IoHT}& \multirow{4}{*}{\makecell[l]{Enables remote patient monitoring; \\ Improves medical diagnosis and treatment; \\ Increases the efficiency of healthcare services} }& \multirow{4}{*}{\makecell{Privacy, \\ Heterogeneity} }\\
    & & \\
    & & \\
    & & \\
    \hline
    \end{tabular}
    \label{application scenarios}
\end{table*}

Table~\ref{application scenarios} summarizes the features of these four application scenarios and the corresponding challenges that can benefit from BlockFL.


\section{Security and Privacy of BlockFL for IoT}
\label{Security and Privacy}

Security and privacy are crucial elements when it comes to FL and Blockchain technology, and this importance carries over to BlockFL as well, making them areas of significant interest and concern.
This section presents an analysis and comparison of BlockFL models from various application domains with a focus on security and privacy. 
Compared to traditional FL, the integration of Blockchain offers BlockFL a stronger and more scalable solution to support security and privacy protection without depending on any centralized server.

\subsection{PIoT}

As PIoT applications become more widespread, the massive amounts of sensitive information used for training models pose significant challenges to privacy protection. 
In recent years, data security and privacy protection have garnered increased attention from researchers, particularly in relation to data generated during the sensing, communication, and computation processes of PIoT.

FL is at risk of data leakage when facing adversaries with an honest-but-curious server~\cite{13} or with Generative Adversarial Network (GAN) technology~\cite{14}. 
Although Blockchain can promote the development of decentralized and data-intensive applications~\cite{15}, FL still relies on the honesty of miners as all raw data are public. 
Therefore, traditional FL and separate Blockchain technologies cannot satisfy the security and privacy requirements of PIoT scenarios.
Hence, the BlockFL, which combines the advantages of FL and Blockchain, has become a new research direction to solve security and privacy issues in the PIoT.

A number of researchers have proposed different solutions for addressing the challenges of security and privacy in FL with the integration of Blockchain technology. 
Awan et al.~\cite{19} present a Blockchain-based PPFL model, which combines the FL framework with the decentralized trust of Blockchain to ensure privacy preservation. 
To achieve this, the authors enhance a variant of the Paillier cryptosystem to implement homomorphic encryption. 
Yin et al.~\cite{20} propose an FDC framework based on FL and Blockchain, which leverages multiparty secure computation technologies to ensure data security. 
Wang et al.~\cite{21} discuss the Security Parameter Aggregation Mechanisms in detail in their BlockFedML model. 
Furthermore, Ma et al. \cite{22} propose a new group-based Shapley value computation framework that is compatible with secure aggregation in a Blockchain-based FL model. 
The approaches aim to address the privacy and security concerns in FL by integrating Blockchain technology and novel cryptographic methods.

\subsection{IIoT}

The proliferation of IIoT has resulted in an exponential increase in the volume of data generated by devices equipped by various industries. 
The value of the data, because of the sensitive information it contains, has gained rise to concerns about data security. 
The leakage of IIoT data could result in significant financial losses for the company, as well as disruption and disorder within the industry.

Ensuring data security is a crucial factor in determining the utility of the IIoT model. 
Wang et al.~\cite{53} identify the security requirements for IIoT and investigate the advantages of integrating Blockchain technology into IIoT applications. 
In a separate study, Blockchain is leveraged in edge intelligence to optimize resource allocation in IIoT~\cite{54}. 
Additionally, FL, a privacy-preserving technique that avoids the sharing of raw data, has also been highlighted in IIoT applications. 

To satisfy differential privacy, Geyer et al.~\cite{55} proposed a method to conceal the contribution of each client during the training process.
In the pursuit of safer data sharing, Lu et al.~\cite{56} build data models with BlockFL structures, where only FL-generated data models are shared by Blockchain. 
And thus, the model reduces the risk of raw data leakage and effectively protects data security. 
By using homomorphic encryption and secure multi-party computation, the authors ensure that the privacy of the raw data is maintained while enabling collaborative learning.
Furthermore, Yazdinejad et al.~\cite{yazdinejad2022block} developed a block hunter framework based on cluster detection to automatically search for attacks and threat risks in BlockFL networks.

\subsection{IoV}

In IoV, practical models require large amounts of data sharing, which can include sensitive private information such as frequently visited addresses, real-time road conditions, driving routes, and driving preferences. 
Protecting privacy while participating in model training and sharing information with others is important and necessary in IoV applications~\cite{zha2017analytic}, where the BlockFL framework could play an influential role. 

Liu et al.~\cite{74} improve an optimized mask noise model upload algorithm for secure secret sharing of model parameters. 
The authors also introduce a two-stage intrusion detection system (IDS) utilizing the combination of FL and Blockchain in vehicles and roadside units to ensure data security and privacy protection in IoV. 
Chen et al.~\cite{75} propose a novel Byzantine-fault-tolerant Blockchain-based FL method named BDFL, which implements a publicly verifiable secret sharing scheme to address privacy concerns in IoV. 
The experimental results on actual datasets demonstrate the practicality of multi-object recognition while preserving privacy.

\subsection{IoHT}

In the IoHT, a large amount of sensitive information poses a significant risk of privacy breaches. To address the issue and enable secure data sharing, BlockFL has been introduced as a promising approach for IoHT applications.
Passerat-Palmbach et al.~\cite{80} have proposed a basic structure of Blockchain-orchestrated FL and identified six critical elements of privacy and security requirements in IoHT models:
\begin{itemize}
    \item Ensuring data security sharing and processing in the Blockchain while maintaining privacy;
    \item Refusing to generate data or fabricate value effectively;
    \item Ensuring computation with FL and advanced cryptography;
    \item Ensuring privacy through both software and hardware cryptography;
    \item Establishing a suitable incentive mechanism to evaluate data quality;
    \item Preventing poisoning attacks and mitigating the impact of poor data.
\end{itemize} 
The requirements provide a comprehensive framework for developing secure and privacy-preserving IoHT applications using Blockchain-orchestrated FL.

Based on the requirements, Połap et al.~\cite{81} develop a multi-agent system that divides specific medical tasks into agent units for parallel training of classifiers with FL, and uses Blockchain to share and protect private data. 
Similarly, Aich et al. \cite{82} design a BlockFL-based solution for the secure sharing of healthcare data to address the fragmented nature of personal medical data. 
However, the approach has only been theoretically analyzed without using practical applications yet.

El Rifai et al.~\cite{83} conduct experiments on a diabetes dataset to evaluate the effectiveness of their model, which utilizes BlockFL for secure knowledge sharing between medical centers while preventing attackers from accessing patients' raw records. 
The experimental results demonstrate the proposed approach's ability to ensure data security and privacy protection. 
A hybrid Blockchain-based FL framework~\cite{84} has been tested in the context of COVID-19 clinical trials, which ensures the complete privacy of training data and supports reputation management, making it more relevant to current healthcare applications.
Another privacy-preservation framework proposed by Singh et al.~\cite{singh2022framework} also illustrated that the BlockFL technology can mitigate the risk of exposing patient medical data, creating a transparent and secure environment for data sharing and model training

\begin{figure}[!ht]
        \centering
        \includegraphics[width=0.7\columnwidth]{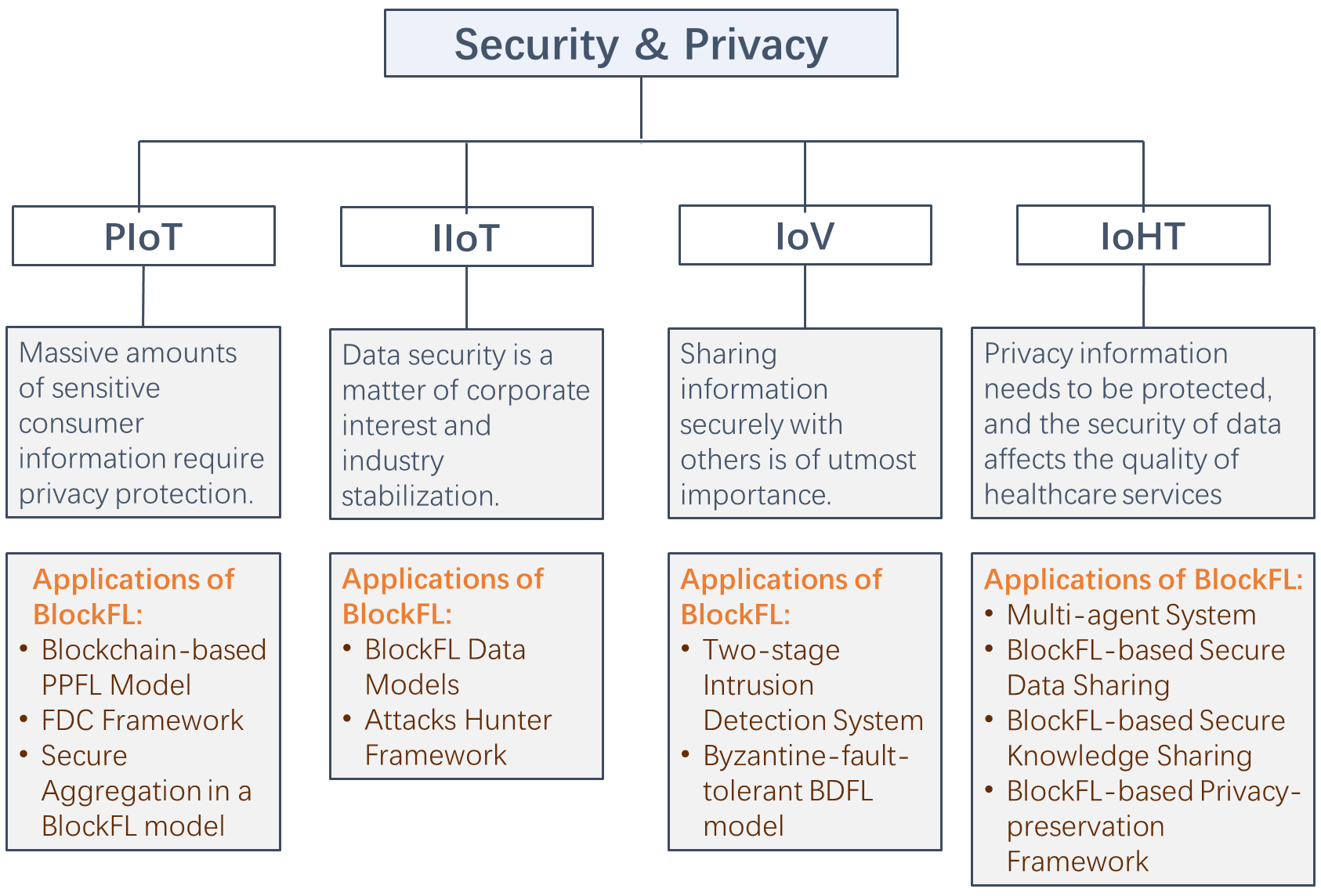}
        \caption{Application of BlockFL: Security and Privacy. The models applied in PIoT and IoHT place greater emphasis on privacy protection, while those used in IIoT and IoV prioritize security. This reflects the different priorities and concerns of each domain, highlighting the need for tailored approaches to security and privacy protection in BlockFL.}
        \label{fig_Security and Privacy}
\end{figure}

Fig.~\ref{fig_Security and Privacy} presents a summary of the latest BlockFL models in various application scenarios, highlighting the importance of addressing security and privacy concerns.

\section{Trust and Reliability of BlockFL for IoT}
\label{Trust and Reliability}

Apart from the improved security and privacy protection, the characteristics of immutability and auditability in Blockchain provide a sense of trust and reliability to the BlockFL process.
In this section, we delve into the ways in which BlockFL addresses the trust and reliability problems and examine the specific features of BlockFL-based models that are tailored to meet the unique needs and characteristics of various application domains.

\subsection{PIoT}

The PIoT model has to ensure the trust and availability of information, and services to enable its practical application. 
To avoid errors and losses in data transmission and communication, which can be catastrophic consequences (e.g., the risk of poisoning attacks), \cite{23} has demonstrated the possibility of malicious workers manipulating model results by injecting poisonous data or tampering with training data.

Existing studies have been developed to tackle the challenge of information errors and losses during traditional FL training. 
Zhao et al.~\cite{24} design a method to address the issue by minimizing the influence of low-quality participants. VerifyNet~\cite{25} introduces a verifiable framework that verifies the integrity of the aggregated results of FL. 
However, the framework faces the problem of susceptibility to single-point attacks due to its reliance on the central server.

The combination of Blockchain and FL technologies improves the reliability of PIoT models because of the auditability and decentralization provided by Blockchain. 
Peng et al.~\cite{26} improve the VFChain system, enabling verification for the FL training process and recording verifiable proofs in the Blockchain. 
Preuveneers et al.~\cite{27} introduce an anomaly detection model into the FL process and utilize Blockchain technology to record its incremental updates. 
Kang et al.~\cite{28} propose a metric called ``reputation" to support reliable-worker selection, ensuring data integrity and preventing tampering. 
By reducing the impact of adversarial data corruption, integrating Blockchain and FL technologies can improve the robustness and stability of PIoT models. 
The work in~\cite{29} applies the concept of reputation in a Blockchain-based fine-grained FL model to facilitate trustworthy collaborative training.
The experiment in~\cite{30} shows that implementing the Blockchain in FL improves the performance when adopting various types of corruption to the end-point adversary's dataset, including salt and pepper noise and circle occlusion.
The FLchain scheme developed by Majeed et al.~\cite{31} outperforms traditional FL models in robustness as the provenance of data is auditable.

\subsection{IIoT}

Real-world industries require reliable and stable IIoT models that can withstand environmental disturbances and attacks. 
However, data flaws are common in the FL process as local datasets are easily disturbed by environmental factors~\cite{57}. 
The usage of Blockchain for the underlying mechanism of the IIoT model can ensure the regular operation of the entire system so that the machines can work honestly and normally~\cite{58}.

To detect device failures and attacks in IIoT applications, Zhang et al.~\cite{59} introduce an optimization of an averaging algorithm called CDW-FedAvg that calculates the distance between positive and negative class data. 
By combining the advantages of both FL and Blockchain, the developed approach reduces the impact of device failures and improves the stability of the IIoT system. 
Similarly, Qu et al.~\cite{60} develop a D2C paradigm for the IIoT model and a modified Markovian decision process to enhance performance when facing poisoning attacks.

Stability is a crucial advantage of industrial automation, enabling control and prediction of the operational status of machines. 
In Industry 4.0, researchers aim to enhance the stability of IIoT models by combining FL and Blockchain technologies. 
Hua et al. \cite{61} conduct experiments on heavy haul rail applications, replacing manual operation with intelligent control using a Blockchain-based asynchronous FL system. 
The simulation results demonstrated that the proposed BlockFL system effectively achieves stable and smart control in real heavy-haul rail applications.

\subsection{IoV}

\smallskip
\noindent\textbf{Timeliness.}
The high mobility of vehicles in IoV introduces dynamically and rapidly changing environments, leading to crucial timeliness of decisions, as autonomous driving and intelligent transportation systems require vehicles to respond quickly to real-world situations. 
Therefore, improving the speed of vehicles' model learning and information communication is a key issue in practical applications and a prominent topic in research.

To accelerate model learning and information communication in the IoV, Pokhrel et al.~\cite{70} develop a mathematical analysis to identify the delay in the BlockFL model, where participating vehicles share their on-vehicle machine learning model updates via Blockchain and cooperate to complete the FL process. 
The vehicles calculate the total end-to-end latency, including communication and consensus delays, and proposed an online algorithm to adjust parameters in real-time to minimize the model delay.

\smallskip
\noindent\textbf{Trust.}
To improve the intelligence of vehicles, different models are needed to process, analyze, and respond to different application scenarios.
Each model in the IoV requires diverse and vast data that is collected from the vehicles, the neighbors, and the roadside units. 
Due to the variability of the IoV, neighboring vehicles are in a constant state of flux, so vehicles in the IoV are often unfamiliar with their surroundings. 
It is, therefore, essential to evaluate the credibility of the data and identify any malicious attempts to compromise it.

Blockchain has been seen as an effective tool to integrate with FL as BlockFL to manage participants and improve system reliability~\cite{71} to address the trustworthiness issue in the IoV. 
PermiDAG model developed by Lu et al.~\cite{72} uses a hybrid Blockchain with directed acyclic graph technology to perform asynchronous FL. 
The quality of shared parameters is verified to detect false information and malicious data. 
Kang et al.~\cite{73} introduce a reputation system to judge participant trustworthiness and design a distributed reputation calculation scheme for selecting trustworthy participants. 
The authors present a stable many-to-one matching model for task assignment to achieve a trusted win-win situation.

\begin{figure}[!ht]
        \centering
        \includegraphics[width=0.65\columnwidth]{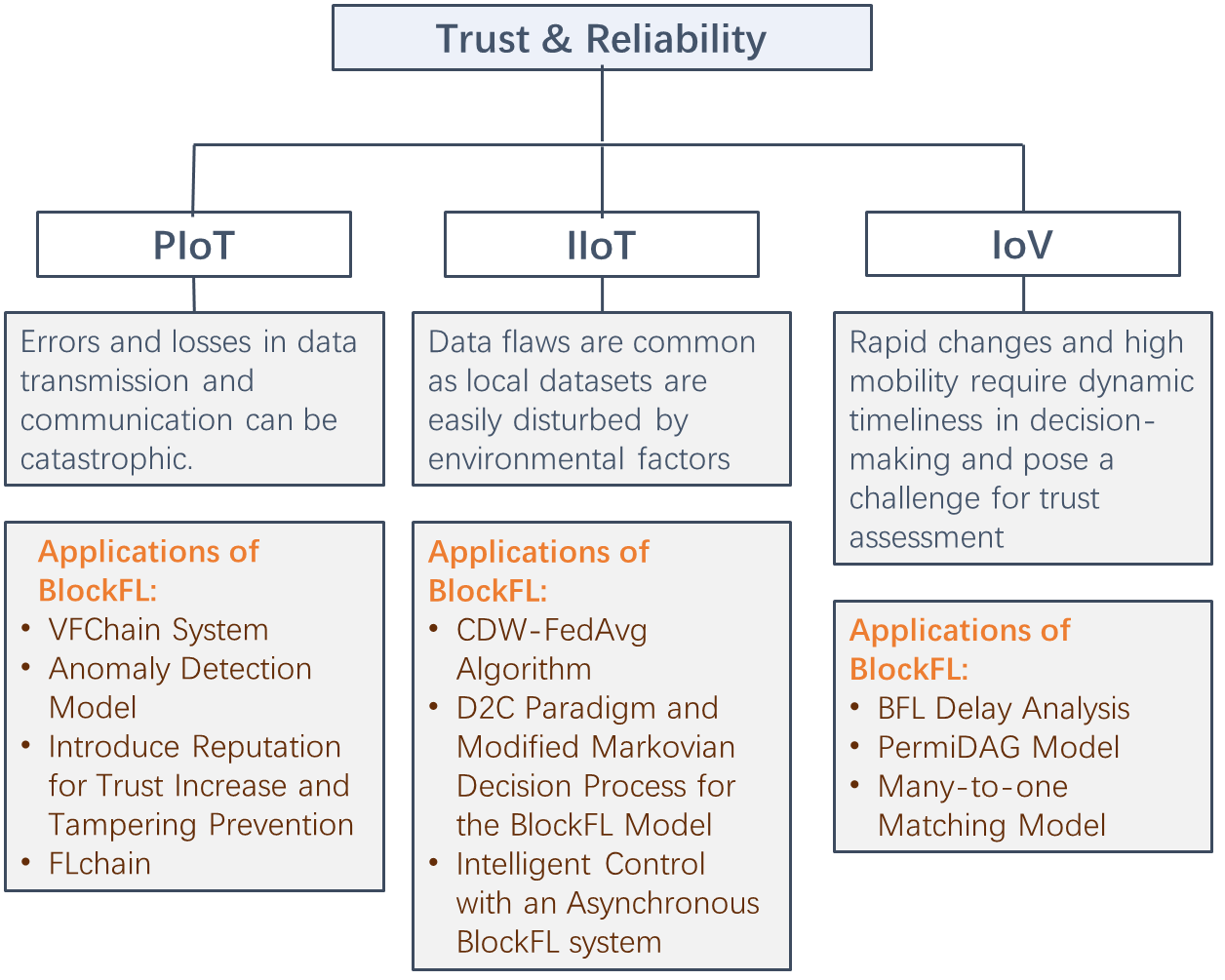}
        \caption{Application of BlockFL: Trust and Reliability. The high-speed changing environment of IoV requires application models to have higher requirements for trust and stability. At the same time, models in PIoT and IIoT also have specific demands for trust and stability.}
        \label{fig_Trusty and Reliability}
\end{figure}

The state-of-the-art models of BlockFL for trust and reliability enhancement are summarized in Fig.~\ref{fig_Trusty and Reliability} to demonstrate its relevance and usefulness in different application domains.

\section{Efficiency of BlockFL for IoT}
\label{Resource Limitation}

In practical application scenarios, realistic conditions such as limited resources and restricted costs must be taken into account, unlike in theoretical analysis. 
As a result, adjusting various factors and seeking efficient solutions under limited conditions is a crucial topic in the development of BlockFL. 
Because BlockFL comprises the resource-intensive learning process of FL and the block generation process of Blockchain, balancing these two parts to achieve optimal system performance requires careful consideration of multiple factors.
This section focuses on the efforts of BlockFL to balance efficiency within resource constraints in various application domains through analysis and discussion.

\subsection{PIoT}
The practicality of the PIoT model is to enable intelligent PIoT devices and offer advanced services. 
To accomplish this ambition, the PIoT model should exhibit high accuracy and efficiency when executing tasks, which means optimizing algorithms and designing models for improving performance under the condition of limited resources.
 
Accuracy is a key metric for model evaluation in PIoT models. 
Existing works have demonstrated that BlockFL models integrating Blockchain and FL technologies outperform traditional FL and separate Blockchain in various tasks. 
Liu et al.~\cite{32} proposed the FedAC model, which combines asynchronous FL and Blockchain technologies, and achieved impressive accuracy rates of 98.96\% in horizontal data distribution and 95.84\% in vertical data distribution, outperforming the accuracy of its counterparts.

Efficiency is also an important metric for evaluating BlockFL models. 
To improve efficiency, Ramanan et al.~\cite{33} present the BAFFLE model by using smart contracts in Blockchain to coordinate the round delineation, model aggregation, and update tasks in FL. 
The model significantly reduces the computational cost of the model because smart contracts are computerized transaction protocols~\cite{34} that automatically execute the contractual terms. 
Feng et al.~\cite{35} develop two complementary policies to ensure efficiency, i.e., controlling the block generation rate and dynamically adjusting the number of training times in asynchronous FL. 
In ChainsFL~\cite{36}, synchronous and asynchronous training are combined to improve the efficiency of the model.

\subsection{IIoT}

In research and analysis, the existing studies assume that devices participating in model training have unlimited energy and ample computing power. 
However, in the real world, industrial machines used in manufacturing often fall short of theoretical ideals. 
Due to the cost considerations, such equipment is subject to constraints on capacity, energy, communication ability, and other aspects. 
For instance, machines with limited computing power require more time to train and update the model, while those with poor wireless channel conditions take longer to transmit information. 
Therefore, it is essential to flexibly adjust model parameters based on actual conditions and enhance model performance under resource limitations.

To address the problem of resource constraints, Nishio et al.~\cite{62} develop the FedCS model, which selects suitable training participants. 
By excluding unqualified machines, as many participants as possible can join the training process under limited conditions, making it suitable for actual industrial applications. 
In addition to participant selection, adjusting other model parameters is also effective. 
Qu et al.~\cite{63} consider a range of factors, including communication, delays, and computation cost, to determine the optimal block generation rate in FL-Block, an autonomous FL system based on Blockchain. 

Reducing energy consumption can be utilized to address the resource issues in BlockFL training. 
Lu et al.~\cite{65} improve a compression technique to reduce communication costs without sacrificing performance. 
The authors consider the instability and complexity of the network connections in the IIoT model, allowing machines to join or leave the training process more freely. 
Kang et al. \cite{66} employ a gradient compression scheme to replace complete gradients with sparse but important gradients, effectively reducing communication overhead.

\subsection{IoV}

The dynamic nature of vehicular networks introduces a challenge in resource allocation. 
Despite the advancements in in-vehicle computing and communication technologies, there still exists a gap in achieving optimal solutions for model learning in theory.
This is especially true in the case of BlockFL, where vehicles need to conduct multiple rounds of communication and require high computational power. 
Hence, exploring ways to adjust parameters effectively to meet the requirements under limited resources is an important research direction in IoV.

Chai et al.~\cite{76} improve a hierarchical FL algorithm that leverages Blockchain technology to include multiple ground chains and one top chain, resulting in reduced computation and sharing consumption. 
The experimental results show the effectiveness of the hierarchical structure. 
Pokhrel et al.~\cite{77} introduce a Blockchain-empowered FL system for drones in 6G networks aimed at disaster response systems. 
The authors focus on the impact of transmission parameters such as power and the number of miners on energy consumption through modeling and simulation that offer valuable insights and potential research directions for future work in this field.
The negative impact of the energy limitation problem of drones on the service time is also discussed in a data collection BlockFL scheme~\cite{9714496}.

\begin{figure}[!ht]
        \centering
        \includegraphics[width=0.65\columnwidth]{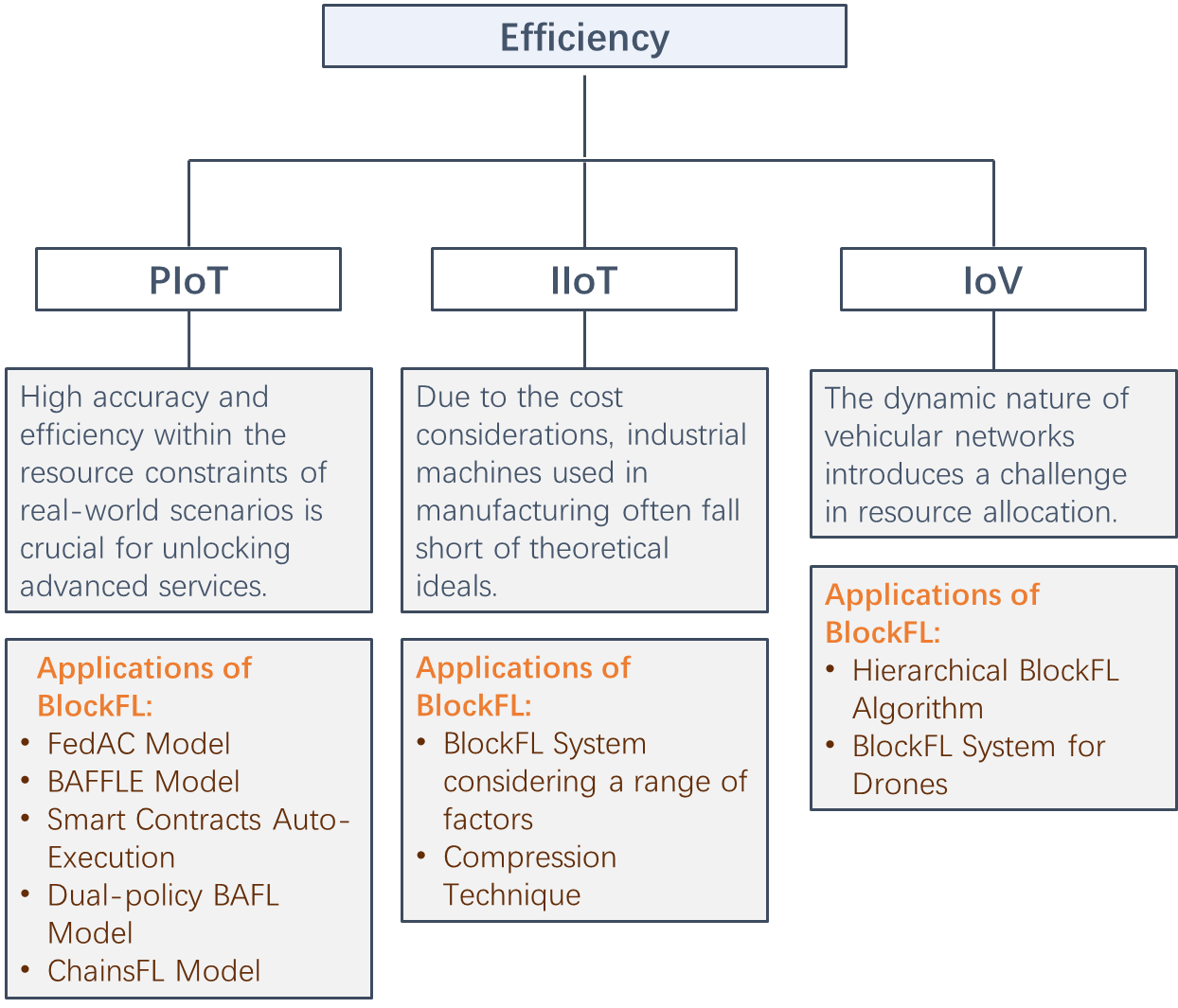}
        \caption{Application of BlockFL: Efficiency. In PIoT, IIoT, and IoV, a variety of resource and economic factors may limit the implementation of BlockFL models. Therefore, it is crucial for application models, particularly those used in IIoT, to consider the balance and optimization between model performance and resources for efficiency.}
        \label{fig_Resource Limitation}
\end{figure}

In Fig.~\ref{fig_Resource Limitation}, we showcase the development of BlockFL in response to resource limitations and unique needs in different application areas.

\section{Data Heterogeneity of BlockFL for IoT}
\label{Measurement Optimization}

The quality and quantity of training data are crucial factors determining the performance of any data-driven ML model, making the expansion of training data richness and diversity an essential issue that cannot be overlooked in the process of model optimization, including BlockFL.
This section emphasizes the efforts made by BlockFL to increase the diversity of training data for performance improvement. 
These efforts include establishing an effective incentive mechanism to encourage more participants in model training, as well as addressing heterogeneous problems to enhance the model's capabilities.

\subsection{PIoT}
Accurate judgments and correct decisions rely on the large amount and diversity of data. So increasing the enthusiasm of devices for participation in the model training is an effective action for optimizing PIoT models' accuracy, which requires the model to have a reasonable incentive mechanism.

The Blockchain has shown its ability to provide an incentive mechanism based on participants' performance effectively. So in recent years, there has been more and more research work to implement Blockchain into PIoT applications, especially with the FL that can safely combine massive devices to train a model together. Without the assumption of honest participants, Short et al. \cite{37} offer rewards on a Blockchain network according to the quality of contributions in FL. And Martinez et al. \cite{38} propose an in-depth workflow to record and reward the contributions of participants. In the work of Kim et al. \cite{39}, Blockchain is used to separate participating users as nodes and induce them to join the FL efficiently.

Besides, in order to attract more participants to join BlockFL to improve data diversity and model performance, more work is devoted to designing more reasonable and attractive incentive mechanisms.
An effective incentive mechanism combining reputation management with smart contracts is proposed by Kang et al. \cite{40} to motivate high-quality devices to join the model learning process. Zhao et al. \cite{41} design an incentive mechanism to award participants with a novel normalization technique. Weng et al. \cite{42} propose a DeepChain framework with a value-driven incentive mechanism to force the participants to train the model following the rules. Kumar et al. \cite{43} also develop a value-driven incentive mechanism to encourage the contributors' positive actions by introducing Blockchain technology via Ethereum.

Introducing repeated competition for FL is also feasible \cite{44} as rational participants want to maximize their profits. Also, based on the hypothesis of rational man, Xuan et al. \cite{45} propose a double-layer FL platform based on Blockchain with an incentive mechanism to ensure that rational workers can gain the maximum benefit by remaining honest. Desai et al. \cite{46} create a general Blockchain-based FL framework to detect and punish attackers automatically. And an honest trainer \cite{47} is presented to gain fairly partitioned profit, rewarding contributions and punishing the malicious. 

\begin{figure}[!ht]
        \centering
        \includegraphics[width=0.65\columnwidth]{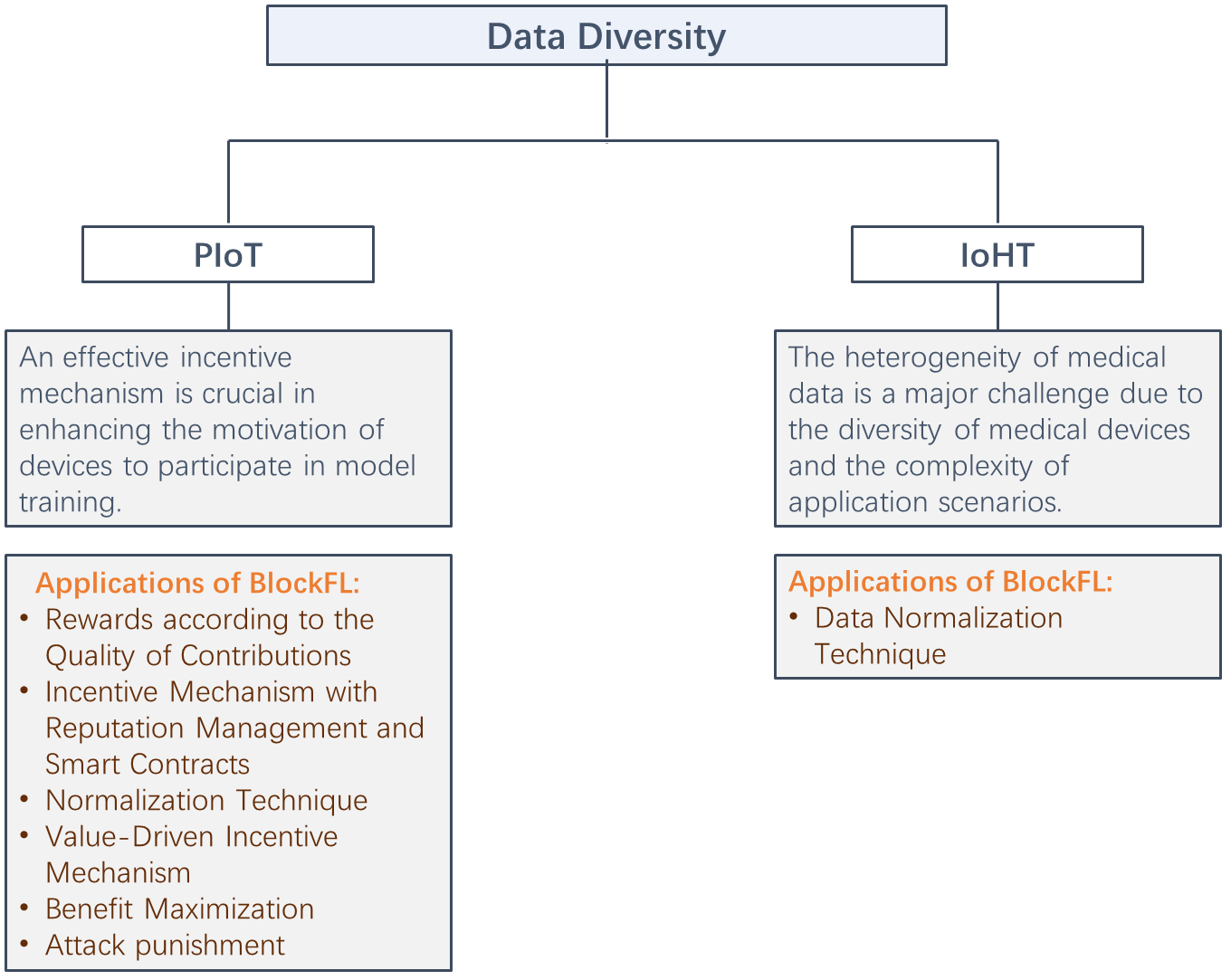}
        \caption{Application of BlockFL: Data Diversity. In order to increase data diversity and optimize BlockFL, the PIoT model should also consider incentives to encourage greater participation, while the IoHT domain presents a challenge with regard to heterogeneous problems that require more attention from the model.}
        \label{fig_Measurement Optimization}
\end{figure}

\subsection{IoHT}
When discussing how to enrich the diversity of training data, the heterogeneity of healthcare data is a significant issue in the context of IoHT due to the diversity of medical equipment and the complexity of application scenarios.
Because of the variability in data format and characteristics across different medical institutions, it is impractical to enforce a standardized structure for all data in IoHT. 
For example, Computed Tomography (CT) images may vary in size, pixel density, and data format, so addressing the heterogeneity of healthcare data is an important consideration for developing effective IoHT solutions.

To address the challenge of heterogeneity in healthcare data, Kumar et al.~\cite{85} improve a Blockchain-empowered FL model with the data normalization technique. 
The authors utilize capsule-network-based segmentation and classification to detect patterns of COVID-19 from various types of lung CT scans. 
By leveraging the Blockchain and FL technologies, the presented BlockFL model caters to the particularities of the IoHT.

The measurement optimization of BlockFL in the direction of PIoT and IHoT is summarized in Fig.~\ref{fig_Measurement Optimization}.


\section{Lesson Learned and Open Challenges of BlockFL for IoT}
\label{Challenges}

The combination of FL and Blockchain has demonstrated significant potential in advancing next-generation digital developments. 
Through theoretical analysis and related experiments, existing studies confirm the application value of integrating BlockFL technologies in various fields. But the works are limited and remain largely theoretical. 
The throughout review of the BlockFL in IoT reveals several essential challenges and unresolved issues when considering the implementation and development of BlockFL.

This section will highlight potential future research directions for BlockFL, exploring both general open issues faced across all application domains, as well as domain-specific challenges.


\subsection{General challenges of BlockFL in Different Application Scenarios}

\begin{table*}[!htbp]
\setlength\tabcolsep{3pt}  
    \centering
    \renewcommand\arraystretch{0.92}   
    \caption{Potential Synergies between BlockFL and Related Technologies}
    \begin{tabular}{|c|c|c|c|c|}
    \hline
    \multirow{5}{*}{\textbf{Technology}}  & \multirow{5}{*}{\textbf{Description}} & \multirow{5}{2cm}{\textbf{Challenges Solved by Combining with BlockFL}} & \multirow{5}{*}{\textbf{Exampl}e} & \multirow{5}{*}{\textbf{Features \& Benefits}} \\
    & & & & \\
    & & & & \\
    & & & & \\
    & & & & \\
    \hline
    \multirow{11}{*}{Cryptography} &  \multirow{11}{3cm}{Mathematical algorithms for security and sensitive information protection.} & \multirow{11}{*}{\makecell{Data Security\\ \& Privacy}} & \multirow{4}{2cm}{PPFL Model~\cite{19}} &  \multirow{4}{4.8cm}{Enhance a variant of the Paillier cryptosystem to implement homomorphic encryption.}\\
    & & & & \\
    & & & & \\
    & & & & \\
    \cline{4-5}
	~ & ~ & ~ &  \multirow{4}{2cm}{FDC Framework~\cite{20}} & \multirow{4}{4.8cm}{Leverages multiparty secure computation technologies to ensure data security.}\\
     & & & & \\
     & & & & \\
     & & & & \\
    \cline{4-5}
	~ & ~ & ~ &  \multirow{3}{2cm}{Fed-DFE Model~\cite{sun2022fed}} & \multirow{3}{4.8cm}{Uses the interactive key generation algorithm to avoid collusion attack.}\\
    & & & & \\
    & & & & \\
    \hline
    \multirow{12}{*}{\makecell{Anomaly\\Detection}} &  \multirow{12}{3cm}{Techniques to identify unusual patterns or outliers in data.} & \multirow{12}{*}{\makecell{Data Security \\ \& Privacy, \\ \\Reliability}} & \multirow{4}{2cm}{Block Hunter Framework~\cite{yazdinejad2022block}} &  \multirow{4}{4.8cm}{Based on cluster detection to automatically search for attacks and threat risks.}\\
    & & & & \\
    & & & & \\
    & & & & \\
   \cline{4-5}
        ~ & ~ & ~ &  \multirow{4}{2cm}{Anomaly Detection Model~ \cite{27}} & \multirow{4}{4.8cm}{Utilizes Blockchain to record the incremental updates of anomaly detection.}\\
    & & & & \\
    & & & & \\
    & & & & \\
    \cline{4-5}
        ~ & ~ & ~ &  \multirow{4}{2cm}{CDW-FedAvg Algorithm\cite{59}} & \multirow{4}{4.8cm}{Calculates the distance between positive and negative class data to detect failures and attacks.}\\
    & & & & \\
    & & & & \\
    & & & & \\
    \hline
     \multirow{13}{*}{\makecell{Optimization}} &  \multirow{13}{3cm}{Methods for finding the best solution for a given problem and \\conditions.} & \multirow{13}{*}{\makecell{Resource \\Limitation}} & \multirow{4}{2cm}{Dual-policy BAFL Model~\cite{35} } & \multirow{4}{4.8cm}{Develops two complementary policies to control block-generation rate and adjust training rounds.}\\
    & & & & \\
    & & & & \\
    & & & & \\
    \cline{4-5}
	~ & ~ & ~ & \multirow{4}{2cm}{FL-Block Scheme~\cite{63} }&\multirow{4}{4.8cm}{Considers delays, communication and computation cost to determine the optimal block generation rate.}\\
    & & & & \\
    & & & & \\
    & & & & \\
    \cline{4-5}
	~ & ~ & ~ & \multirow{5}{2cm}{Disaster ResponseSystem~\cite{77}} &\multirow{5}{4.8cm}{Discusses the effect of the number of miners, computing power, transmission capacity, and channel dynamics.}\\
    & & & & \\
    & & & & \\
    & & & & \\
    & & & & \\
    \hline
     \multirow{7}{*}{\makecell{Compression \\Technique}} &  \multirow{7}{3cm}{Techniques for reducing data size while maintaining integrity and usefulness.} & \multirow{7}{*}{\makecell{Resource \\Limitation,\\ \\Scalability}} & \multirow{3}{2cm}{PAFLM~\cite{65}} &\multirow{3}{4.8cm}{Reduces communication costs without sacrificing performance.}\\
    & & & & \\
    & & & & \\
    \cline{4-5}
	~ & ~ & ~ & \multirow{4}{2cm}{Decentralized FEL Model~\cite{66}} &\multirow{4}{4.8cm}{Replaces complete gradients with sparse but important gradients to reduce communication overhead. }\\
    & & & & \\
    & & & & \\
    & & & & \\
    \hline
    \multirow{4}{*}{\makecell{Data \\Normalization}} &  \multirow{4}{3cm}{Techniques for transforming data into a standardized form.} & \multirow{4}{*}{Heterogeneity} & \multirow{4}{2cm}{Pathological Detection~\cite{85}} &\multirow{4}{4.8cm}{Deals with the data collected by different kinds of CT scanners effectively.}\\
    & & & & \\
    & & & & \\
    & & & & \\
    \hline
    \end{tabular}
    \label{tab:lesson_label}
\end{table*}

For all IoT application scenarios, the promotion and development of BlockFL as an effective distributed model learning solution require the provision of secure data and model protection, economically feasible and scalable schemes, as well as the capability to handle diverse data and multiple tasks.

\begin{itemize}
    \item Data Security \& Privacy.
    Despite the ongoing research efforts, the existing researches are limited and preliminary, so flexible models and systems are required to effectively protect data privacy during the model learning process. 
    The security challenges encountered by BlockFL are continuously evolving, given the revolutionary changes in the way attacks are executed and the iterations of attack models.
    BlockFL is required to respond with different attempts to cope with the new adversary assumptions, as some of the currently widely used assumptions may become outdated and require constant updating.
    Furthermore, the general public's awareness and expectations regarding privacy are constantly rising, resulting in new privacy protection requirements.
    Therefore, BlockFL's optimization in the direction of security and privacy protection must keep up with the times, leaving much room for further research and innovation.
    
    \item Resource Limitation \& Scalability.
    Scalability is a crucial factor in commercializing BlockFL models. 
    As the number of digital devices and data volume increases rapidly, a scalable system is required to handle the growth in complexity. 
    Traditional FL and Blockchain are designed for multiparty cooperation, making them suitable for elastic systems. 
    However, there is still a shortage of related research in this area, and more efforts are needed to address scalability issues with practical considerations. 
    It is also necessary to consider whether modules such as the consensus algorithm in Blockchain can adapt to the limited battery capacity and computing power of many devices in actual application scenarios. 
    Therefore, lightweight computing and low-latency solutions under restrictive conditions are also crucial research directions for the future of BlockFL.

    \item Heterogeneity.
    With the increase in the number of devices increases and the variety of data types, it is necessary to standardize the data for efficient processing. 
    Data loss and errors caused by data acquisition and transmission are significant challenges that must be addressed. 
    Innovative normalization techniques are required to address the challenges associated with processing heterogeneous data, which necessitates a deep understanding of data structures. 
    Intelligent data processing not only enhances the performance of BlockFL models but also stimulates further research on data-driven models.
    
\end{itemize}

Table~\ref{tab:lesson_label} summarizes the relevant technologies that have great potential to drive the future development of BlockFL. 
In consideration of security enhancement and privacy protection, implementing encryption and secure computing technologies can enhance the security of BlockFL models. 
Some attempts to combine blockFL and cryptography~\cite{19,20,sun2022fed} have been verified to be effective for strengthening security.
By leveraging a combination of different encryption algorithms, noise addition methods~\cite{yu2022dataset,yuan2023amplitude} or multi-party security technologies, the development of Blockchain-based FL models can be further improved, and new breakthroughs can be achieved.
Some other works have suggested that BlockFL can address resource limitations and data heterogeneity problems by leveraging data processing techniques, such as compression~\cite{65,66} and normalization~\cite{85}.
These techniques can also be combined with smart contracts~\cite{makhdoom2019privysharing} and sharding mechanisms~\cite{yu2020survey,zhang2022community} to further strengthen BlockFL's autonomy and extend its capabilities.

Additionally, there is a need to explore and conduct further research on a more in-depth combination of Blockchain and FL technologies.
Most of the existing BlockFL models only use Blockchain as a means of aggregation and ledger in the FL process, with a focus on enhancing FL. 
Conversely, there is only a little mention of optimizing and enhancing Blockchain through the FL process in BlockFL.
Some intermediate results in BlockFL, such as the quality of local models, can be used as consensus calculations in Blockchain process~\cite{56}, reducing the cost of computational and communication resources.
Hence, exploring new consensus methods and smart contract technologies based on BlockFL could be a valuable new direction for further development.
At the same time, issues previously discussed only in the blockchain, such as miners' collusion, the possibility of generated blocks being challenged and hybrid-blockchained structures~\cite{yu2021novel,wang2019high}, should also be considered and analyzed in the context of BlockFL.

\subsection{Unique Challenges of BlockFL in Different Application Scenarios}

The integration of BlockFL in various application fields also brings unique challenges due to the distinct characteristics of each area. 
Table~\ref{Unique Challenges} provides an overview of the unique challenges in different fields, their underlying causes, and potential solutions for addressing them.

\begin{table*}[!htb]
\setlength\tabcolsep{3pt} 
    \centering
    \renewcommand\arraystretch{1}   
    \caption{Unique Challenges of BlockFL in different application scenarios}
    \begin{tabular}{|c|c|c|c|}
    \hline
    \multirow{2}{*}{\textbf{Scenario}} &  \multirow{2}{*}{\textbf{Unique Challenges}} & \multirow{2}{*}{\textbf{Reasons}} & \multirow{2}{*}{\textbf{Solutions}}  \\
    & & & \\
    \hline
       \multirow{2}{*}{PIoT}  & \multirow{2}{*}{Collaborative Intelligence} & \multirow{2}{*}{Personalized Needs}  &  \multirow{2}{*}{Transfer Learning Technology} \\
       & & & \\
    \hline
       \multirow{3}{*}{IIoT}  &  \multirow{3}{*}{\makecell{Collaborative Intelligence\\ \& Balancing Costs}} &   \multirow{3}{*}{Industrialization} & \multirow{3}{*}{ Transfer Learning Technology} \\
       & & & \\
       & & & \\
    \hline
       \multirow{3}{*}{IoV}  & \multirow{3}{*}{High-speed stability} & \multirow{3}{*}{\makecell{Real-time Changing \\ Traffic Environments}} & \multirow{3}{*}{Online Learning Algorithms} \\
       & & & \\
       & & & \\
    \hline
       \multirow{2}{*}{IoHT} & \multirow{2}{*}{Permission \&Identity management} & \multirow{2}{*}{High Sensitivity of Medical Data} & \multirow{2}{*}{Consortium Blockchains} \\
       & & & \\
    \hline
    \end{tabular}
    \label{Unique Challenges}
\end{table*}

In large-scale PIoT and IIoT, collaborative intelligence has emerged as a new research direction. 
As existing studies have focused on optimizing a single task, the increasing demand for multitasking collaboration and cooperation requires a complex system to analyze and coordinate the relationship and connection of multitasking and multi-objective. 
To enable intelligent coordination, the models should explore transfer learning technology and other related novel technologies in combination with BlockFL models.
In particular, the process of industrialization requires more consideration of the costs of large-scale implementation.

In IoV, ensuring the stability of the system in high-speed movement is a crucial research direction. 
A large number of vehicles in the IoV application scenarios are constantly moving at high speeds and changing positions in real time, which poses a considerable challenge to the stability and reliability of the network and connection. 
To address this issue, researchers can increase the calculation effectiveness, reduce model delay, and consider optimizing and innovating BlockFL models by imitating online learning algorithms.
Moreover, future vehicles in 6G are expected to support cross-domain communication across the ground, underwater and air~\cite{guo2022vehicular}, so stability in combination with new devices and technologies, such as over-the-air computing, should also be taken into account~\cite{zheng2022balancing}.

In IoHT, permission and identity management of participants are critical challenges due to the high sensitivity of medical data. 
Consortium Blockchains are more suitable for implementation in IoHT, with the high professional knowledge required by participants to analyze and manage medical-related data. 
The involvement of medical organizations can make the management of IoHT models highly controllable and convenient, and multiple participation can improve the accuracy and other performance of IoHT models. 
Thus, researchers should explore how to incentivize participation in BlockFL while considering the problem of membership management.


\section{Conclusion and Future Work}
\label{Conclusion}
In this paper, we have divided the different application scenarios for the conjunction of FL and Blockchain into four important IoT domains: PIoT, IIoT, IoV, and IoHT. We have introduced the status quo and current requirements in each application field and classified the different models according to the solved issues. 
In addition, we have summarized the common challenges in these areas, such as outstanding issues in privacy security, system scalability, and data heterogeneity, and provided several possible future research directions for different fields. 
The specific challenges encountered by BlockFL development in various application domains have been also highlighted, along with some recommendations for further investigation.
Our research has shown that BlockFL, as a highly secure and efficient approach for distributed model training, offers superior performance compared to traditional FL in all IoT application domains thanks to its decentralized structure and transparency.

In our future work, we plan to do further research on optimizing the performance of existing models and improving the practicability of applications. And based on this survey, constructing a new type of Blockchain-based FL model with high privacy security and high accuracy is also a feasible direction for our follow-up works.


\bibliographystyle{ACM-Reference-Format}
\bibliography{sample-base}










\end{document}